\documentclass[pdflatex,sn-mathphys,iicol]{sn-jnl}

\jyear{2022}%

\theoremstyle{thmstyleone}%
\theoremstyle{thmstyletwo}%

\theoremstyle{thmstylethree}%

\usepackage{color,soul}
\usepackage{commath}

\begin{document}

\title[Vehicle Trajectory Prediction on Highways Using Bird Eye View Representations and Deep Learning]{Vehicle Trajectory Prediction on Highways Using Bird Eye View Representations and Deep Learning}


\author[1]{\fnm{Rubén} \sur{Izquierdo}}\email{ruben.izquierdo@uah.es}

\author[1]{\fnm{Álvaro} \sur{Quintanar}}\email{alvaro.quintanar@uah.es}
\equalcont{These authors contributed equally to this work.}

\author[1,2]{\fnm{David} \sur{Fernández Llorca}}\email{david.fernandezl@uah.es}
\equalcont{These authors contributed equally to this work.}

\author[1]{\fnm{Iván} \sur{García Daza}}\email{ivan.garciad@uah.es}
\equalcont{These authors contributed equally to this work.}

\author[1]{\fnm{Noelia} \sur{Hernández}}\email{noelia.hernandez@uah.es}
\equalcont{These authors contributed equally to this work.}

\author[1]{\fnm{Ignacio} \sur{Parra}}\email{ignacio.parra@uah.es}
\equalcont{These authors contributed equally to this work.}

\author[1]{\fnm{Miguel Ángel} \sur{Sotelo}}\email{miguel.sotelo@uah.es}
\equalcont{These authors contributed equally to this work.}

\affil[1]{\orgdiv{Computer Engineering Department}, \orgname{Univeridad de Alcalá}, \country{Spain}}

\affil[2]{\orgname{European Commission, Joint Research Centre (JRC)}, \orgaddress{\city{Sevilla}, \postcode{41092}, \country{Spain}}}


\abstract{This work presents a novel method for predicting vehicle trajectories in highway scenarios using efficient bird's eye view representations and convolutional neural networks. Vehicle positions, motion histories, road configuration, and vehicle interactions are easily included in the prediction model using basic visual representations. The U-net model has been selected as the prediction kernel to generate future visual representations of the scene using an image-to-image regression approach. A method has been implemented to extract vehicle positions from the generated graphical representations to achieve subpixel resolution. The method has been trained and evaluated using the PREVENTION dataset, an on-board sensor dataset. Different network configurations and scene representations have been evaluated. This study found that U-net with 6 depth levels using a linear terminal layer and a Gaussian representation of the vehicles is the best performing configuration. The use of lane markings was found to produce no improvement in prediction performance. The average prediction error is 0.47 and 0.38 meters and the final prediction error is 0.76 and 0.53 meters for longitudinal and lateral coordinates, respectively, for a predicted trajectory length of 2.0 seconds. The prediction error is up to 50\% lower compared to the baseline method.}

\keywords{Trajectory prediction, Highways, Convolutional Neural Networks, Autonomous Vehicles}



\maketitle


\textbf{This work has been accepted for publication at Applied Intelligence.}

\section{Introduction}
\label{sec:introduction}
Autonomous vehicles research has experienced an important growth in the last few years, gradually becoming a reality on the roads at the present time. In this period, robot-vehicles have become more and more sophisticated pieces of hardware capable of sensing the world around them. Simultaneously, limited automated driving functions, such as Automatic Emergency Braking (AEB), Adaptive Cruise Control (ACC), Lane Departure Warning (LDP), and Lane Keeping Assist (LKA) have been successfully released, reaching vehicles sold in the retail market. However, autonomous driving is a completely different and more complex task, including sensing, processing, reasoning, and decision in a wide variety of unexpected scenarios and situations. For these reasons, highways are among the most common driving scenarios in which autonomous vehicles are starting to develop their autonomous capabilities. 
Highway scenarios are structured environments involving vehicles with relatively smooth movements, which contributes to minimize complexity. However, despite the limited complexity, the undefined evolution of the road and vehicle configuration represents some limitations. Decision-making has become one of the most critical tasks in this scenario because of the uncertainties related to the unknown future positions of surrounding vehicles. Thus, accurate predictions of surrounding vehicles' trajectories become of paramount importance as underlying support for the decision-making process and can greatly help to improve comfort and safety in terms of dynamic factors, avoiding sharp maneuvers in highway environments and, ideally, minimizing the use of braking along the route, something that could be achieved if a significant fluidity of traffic could be reached, preventing traffic jams, congestion, and crashes.

The uncertainty associated with the future position of vehicles can be reduced or eliminated by using communication technologies such as Vehicle to Vehicle (V2V), Vehicle to Infrastructure (V2I), WiFi, or 5G to share their trajectories. However, human-driven vehicles cannot share their future positions because they are self-generated in real time and unknown in advance. Unfortunately, automated and human-driven vehicles will share the road for a long period, being prediction skills still necessary to forecast trajectories of human-driven vehicles.

The evolution of Autonomous Vehicles (AVs) in recent years has been boosted by two factors: the development of deep learning models and the increasing computing capabilities of Graphics Processing Units (GPUs). 
Classification tasks usually performed by hand-crafted algorithms are now replaced by deep learning models that learn automatically. The development of Convolutional Neural Networks (CNNs) has played a major role in image processing, enabling image classification, detection, or segmentation. These systems have reached levels of accuracy that sometimes exceed human capability, and can also work in real time thanks to advances in GPU computing. These two factors allow processing not only isolated images, but also video streams in real time to classify actions as time-based events. AVs can take advantage of these algorithms to predict the evolution of traffic scenes and anticipate critical situations.

In this paper we extend our preliminary work \cite{Izquierdo2020} presenting a novel method to predict trajectories in highway scenarios using an efficient Bird Eye View (BEV) representation and a state-of-the-art CNN to predict the future state of the vehicle configuration by performing an image-to-image regression task. The idea is that given an input set of consecutive images the CNN can generate an output set of consecutive images that represents the evolution of the input set.
Vehicle positions and road structure are encoded in a simple but effective schematic BEV representation, virtually unlimited in the number of vehicles and number of lanes.
Each time step is represented in an individual BEV representation. 
A complete sequence of images is stacked to establish the input and output data as the way to incorporate temporal dependencies and modeling vehicle dynamics.
Interactions are implicitly considered as spatial relationships between elements and their temporal evolution. 
In addition, interactions between vehicles are learned as the network extracts patterns from all vehicles and the road configuration simultaneously.
The presented approach is suitable to perform predictions for multiple surrounding vehicles and time horizons in a single computational step. 

In contrast to time-series data approaches, our method does not require a fixed input data structure or a fixed interaction scheme. Spatial-temporal  dependencies and interactions are expressed by graphical representations and are implicitly modeled by the spatial condensation of the CNN.
In addition, the nature of the input and output data is the same, allowing a straightforward interpretation of the results, in accordance with the requirements of explainability and transparency \cite{Llorca2021}.

Based on this interpretable approach and with a relatively simple architecture, we achieve state-of-the-art results on datasets obtained from in-vehicle recording systems as well as from infrastructure or top-view recording systems.

The model proposed in this work contributes to: (1) easy integration of any prediction feature; (2) the interactions between vehicle and features are implicitly modeled based on the special topology of the selected CNN architecture; and (3) the multi-vehicle and multi-horizon prediction is performed in a single-step fashion for an unlimited number of vehicles.

Future application scenarios of the proposed method are in-vehicle prediction systems as safety and comfort enhancers, improving human reaction time and smoothing out abrupt maneuvers. This prediction model, due to its general formulation, can also be used as a traffic management system tool at controlled intersections.

After the introduction section, the state of the art is reviewed in section \ref{sec:arte}, analyzing the most relevant works in trajectory prediction and the available datasets to develop and test them. The problem approach and the description of the proposed model are detailed in section \ref{sec:trajectory}, describing the encoding and decoding information process and the network architecture. This section provides training strategies and parameters, as well as the definition of a baseline method for comparison purposes. Section \ref{sec:trajectory_results} presents quantitative and qualitative trajectory prediction results derived from the developed model. Finally, conclusions and future work are exposed in section \ref{sec:conclusions}.

\section{Related Work} \label{sec:arte}

\setlength{\tabcolsep}{2pt}
\begin{table*}[t]
\footnotesize
\centering
    \renewcommand{\arraystretch}{1.1}
    \begin{tabular}{lc|c|ccc|c|c}
        \multicolumn{2}{c|}{\textbf{Work}} & & \multicolumn{3}{c|}{\textbf{Input}} & & 
        \textbf{Prediction} \\
        \textbf{Authors} & \textbf{Year} & \textbf{Dataset} & \textbf{Kinematics} & \textbf{Context} & \textbf{Interaction} & \textbf{Model} & \textbf{Target}  \\
        \hline
        Hermes et al. \cite{Hermes2009}     & 2009 & Own       & Ego-vehicle                   & -             & -               & ANN-RBF       & Ego-vehicle \\
        Ammoun et al. \cite{ammoun2009}     & 2009 & Own       & Ego-vehicle                   & -             & -               & KF            & Ego-vehicle \\
        Ranjeet et al.\cite{Ranjeet2010}    & 2010 & NGSIM     & Single                        & \checkmark    & -               & NN            & Center\\
        Wiest et al.\cite{Wiest2012}        & 2012 & Own       & Ego-vehicle                   & -             & -               & GMM           & Ego-vehicle \\
        Houenou et al.\cite{Houenou2013}    & 2013 & PKU       & Ego-vehicle                   & -             & All surr.       & CYRA          & Ego-vehicle \\
        Yao et al. \cite{WenYao2013}        & 2013 & PKU       & Ego-vehicle                   & -             & All surr.       & Database      & Ego-vehicle \\
        Yoon et al. \cite{Yoon2016}         & 2016 & NGSIM     & Single                        & -             & -               & ANN           & Center \\
        Izquierdo et al. \cite{Izquierdo2017} & 2017 & PKU     & Ego-vehicle                   & -             & -               & ANN + SVM     & Single\\
        Altché et al. \cite{Altche2017}     & 2017 & NGSIM     & Single                        & \checkmark    & TTC             & LSTM          & Single\\
        Kim et al. \cite{kim2017}           & 2017 & Own       & Ego-vehicle + Surr.           & \checkmark    & 36X21 Grid      & LSTM          & Surrounding \\
        Deo et al. \cite{Deo2018}           & 2018 & NGSIM     & Center + Surr.                & \checkmark    & 3X13 Grid       & LSTM + CSP    & Center\\
        Hu et al. \cite{hu2018}             & 2018 & NGSIM     & Center + Surr.                & \checkmark    & SIMP            & CVAE          & All \\
        Benterki et al. \cite{Benterki2018} & 2018 & NGSIM     & Center                        & \checkmark    & 3X2 Grid        & LSTM-GRU      & Center \\
        Roy et al. \cite{Roy2019}           & 2019 & VISDRONE  & Single                        & \checkmark    & Appearance      & GAN           & Single \\
        Kim et al. \cite{Kim2020}           & 2020 & HighD     & Surrounding                   & \checkmark    & Attention       & Attention     & Surrounding \\
        Khakzar et al. \cite{Khakzar2020}   & 2020 & NGSIM \& HighD & Center + Surr.           & \checkmark    & Spatial + TTC   & LSTM          & All \\
        Song et al. \cite{song2020pip}      & 2020 & NGSIM \& HighD & Center + Surr.           & \checkmark    & Ego-vehicle     & LSTM + CSP    & Surrounding \\
        Lin et al. \cite{Lin2021}           & 2021 & NGSIM     & Center + Surr.                & \checkmark    & 3X13 Grid       & LSTM          & Center \\
        Nejad et al. \cite{Nejad2021}       & 2021 & HighD     & Center + Surr.                & \checkmark    & Appearance      & CNN + CSP + LSTM & Center \\
    \end{tabular}
    \caption{Analisys of Trajectory Prediction Works.}
    \label{tab:traj-arte}
\end{table*}
\setlength{\tabcolsep}{6pt}

This section reviews vehicle trajectory prediction works in highway scenarios by analyzing input data, algorithms, the type of generated prediction and, especially, the prediction target.

Prediction target can be classified into two categories for trajectories recorded from on-board sensors: the ego-vehicle and the surrounding vehicles. The main difference between ego and surrounding vehicles is the accuracy and availability of data in their neighborhood. The ego-vehicle has precise measures of its state and the state of the nearest neighbors. However, the surrounding vehicles' state could be accurate but the estimated state of their neighbors can be degraded or even not available due to the sensors' range limitations or physical occlusions. 
For trajectories recorded from a static and exterior point of view, such as infrastructure acquisition systems or drones, the ego vehicle does not exist and the accuracy and availability of data is the same for all the vehicles. Conceptually, the prediction target definition fits better into the ego-vehicle description, because the accuracy and availability of data is guaranteed for each vehicle and their neighbors, reason for which the problem is simplified.

The trajectory prediction problem addresses the forecasting of one or several future positions of an analyzed vehicle or a group of them. Trajectories denote a set of positions with a corresponding timestamp, but a single predicted position combined with the current position of the vehicle can define a trajectory.
Positions are used to define a precise location, either in 2D or 3D reference systems, in local or global frameworks. Independently of the reference system, positions are described by numbers and commonly trajectory predictions are tackled from numeric approaches. Almost all of the works reviewed are based on variables that describe the motion history of vehicles by representing their state in numerical form, either in a continuous \cite{Hermes2009, ammoun2009, Ranjeet2010, Wiest2012, Houenou2013, WenYao2013, Yoon2016, Altche2017, hu2018, Schorner2019, Benterki2018, Kim2020, Khakzar2020, song2020pip, Nejad2021} or discretized \cite{kim2017, Deo2018, Lin2021} space. Only one approach addressed the trajectory prediction problem from a graphical perspective \cite{Roy2019}, generating predictions directly over images.

The state of the art is reviewed regarding these three categories: data used as input and how it is structured, type of generated data, and databases used to develop the models.

The input variables range from simple position sequences to complex road representations. Based on the nature of the data, they can be classified into three major groups.

Kinematic and dynamic variables such as position, speed, acceleration, heading, and yaw rate define the state vector in a detailed manner. Early works used this vehicle representation (in whole or in part) considering only the prediction target by itself \cite{Hermes2009, ammoun2009, Wiest2012}. These approaches learn simple physical-based motion models that cannot anticipate any maneuver until it has been explicitly observed in the input sequence. Usually Kalman Filter (KF), Gaussian Mixture Models (GMM), Artificial Neural Network (ANN), and Recurrent Neural Network (RNN) are used in these approaches. Although these methods performed properly, they are based on a explicit physical model, so scaling and generalization capabilities are limited. To address these shortcomings, contextual variables were included, such as lateral and longitudinal positions, lateral speed and acceleration, or heading error. These variables represent the combination of the vehicle state and the road parameters. The transformation includes lane information indirectly, which allows models to learn road-based vehicle trajectories \cite{Yoon2016, Altche2017}. 

Vehicles interactions are even more conditioning than road configurations, but their integration can be considered in many different ways. The main problem adding interactions is the varying number of involved vehicles. 
A simple method to include interactions was addressed in \cite{Altche2017}, where the Time To Collision (TTC) is appended to the vehicle state input. TTC represents in a single number if one vehicle is approaching another and arises the need for a lane change or a speed reduction. However, the decision of changing lanes depends on many factors, such as the availability of the adjacent lanes or social agreements (i.e., overtaking is only allowed in one direction in most countries). TTC is also used in \cite{Khakzar2020} to define a risk map around the ego-vehicle. 
In \cite{Benterki2018} a simple vehicle-centric structure is proposed to integrate adjacent vehicle interactions at six tentative positions around the centered prediction target.
A fixed spatial configuration is proposed in \cite{kim2017} to incorporate existing vehicles into the algorithm. The road space is divided into small and equal areas to define a lattice where the state vector (position and speed) of each possible existing vehicle is included together with the ego-vehicle state vector. However, this technique follows a vehicle-centered approach and only models how surrounding vehicles affect the ego-vehicle. In \cite{Deo2018} the same philosophy is adopted, dividing the road area into many small rectangular divisions. The state vector of each vehicle is included in each corresponding division. The difference arises with the use of a so-called \textit{Convolutional Social Pooling} (CSP) block. This block uses spatial connections between the existing divisions and learns motion patterns to assess how the surrounding vehicle configuration affects the prediction target. In contrast with \cite{kim2017}, this approach is vehicle-centered. This method can be applied to any other traffic participants, but the availability or the quality of the measures could change.  
The same spatial grid is applied in \cite{Lin2021} in which multiple LSTMs are combined with spatial–temporal attention mechanisms. Interactions are also modeling in \cite{Kim2020} by using multi-head attention mechanisms. The impact of the future trajectories planned by the ego-vehicle on the future trajectories of the surrounding vehicles can be explicitly included as an input \cite{song2020pip}.
Other proposals, such as \cite{Houenou2013, WenYao2013}, store road configuration examples generating a knowledge database using trajectory stretches and/or the road occupancy configuration.

Vehicles' state, road configuration, interactions, and context are needed for a precise scene understanding and all of them are included under visual appearance in images. In \cite{Roy2019}, video sequences are used to generate future vehicle locations at the image reference system by means of a Generative Adversarial Network (GAN). This approach leads with the limitations to model road configuration, context, or interactions; on the other hand, predictions are limited to the image domain. 
Another approach is to extract features from bird's-eye view images using a CNN model, and obtain the predictions using LSTM-based systems \cite{Nejad2021}. We also highlight work originally designed for pedestrian trajectory prediction, which has been successfully adapted for vehicle trajectories in highways environments, such as Social LSTM \cite{Alahi2016}, Convolutional Social Pooling LSTM \cite{Deo2018_2} or Social GAN \cite{gupta2018social}. 

Prediction models are clearly conditioned by road structure and vehicle configuration. Based on this, different kinds of trajectories can be predicted, attending to the nature of the used algorithms.

Regarding the number of trajectories predicted they could be classified into two categories. The first category is the single-vehicle prediction, where the prediction is focused only on one vehicle and surrounding vehicles actuate as conditioning factors. These approaches need to repeat the prediction process for each existing vehicle. However, despite its limitations, it is the most widely used technique.
The second category is the multi-vehicle predictions, where the trajectories of all the involved vehicles are predicted at the same time.

The datasets used to develop the trajectory prediction works are limited to some private custom datasets and two public datasets. Some works, such as \cite{Hermes2009, Wiest2012, kim2017}, use their own datasets, that are not publicly available. In these cases, the datasets were recorded from on-board sensors. The main public dataset used for trajectory predictions is the NGSIM dataset \cite{NGSIMHW101, NGSIMI80}, which has been used widely \cite{Ranjeet2010, Yoon2016, Altche2017, Deo2018, hu2018, Benterki2018}. A small minority used the PKU dataset \cite{Houenou2013,WenYao2013}, an on-board recorded dataset. The NGSIM is the most used dataset to develop trajectory prediction systems because of its simplicity and antiquity, as it can be observed in table \ref{tab:traj-arte}. This dataset offers precise and non-occluded data from an infrastructure point of view in a straight stretch of highway. Recent works have brought novel datasets trying to fulfill the limitations of the existing datasets. The HighD \cite{highDdataset} is a drone-recorded dataset that provides a massive number of trajectories in German highways. 
The PREVENTION dataset \cite{izquierdo2019prevention} is an on-board recorded dataset including LiDAR, front and back cameras, radars, and an RTK GPS featuring Inertial Navigation System (INS).
Other datasets, non limited to highways, have been released in the last years, such as the Argoverse, Waymo, inD, rounD, and INTERACTION \cite{Argoverse, waymo_open_dataset, inDdataset, rounDdataset, zhan2019interaction} datasets to encourage trajectory predictions on urban areas. 

It is important to note that when data is obtained from on-board sensors, including cameras, it is possible to detect \cite{Yin2021} and track \cite{Wang2022} vehicles from a realistic perspective, and include additional appearance information that is not possible to obtain from sensors on the infrastructure or drones. 

Table \ref{tab:traj-arte} presents a summary of the analyzed works providing key features and distinguishing factors. Note that references to the ego vehicle are related to works that addressed the ego-trajectory prediction problem for on-board sensors. In the case of works based on external sensors (NGSIM), the label \textit{single} is used to denote single trajectories that could be considered equivalent to ego trajectories. Reference to \textit{center} is relative to a vehicle considering all their surrounding vehicles, which are abbreviated with notation \textit{surr}.

\section{Problem Approach}\label{sec:trajectory}
Trajectory prediction addresses the problem of knowing the future position of a vehicle or group of them. 

The proposed method employs a deep learning image-to-image regression approach to forecast the position of the vehicles in a given scene. Conceptually, road configuration and vehicle positions are represented in a schematic BEV image. Then, a set of consecutive representations are stacked to create a simplified video sequence. The prediction core, a CNN, generates the video sequence that follows the corresponding input video sequence.

\subsection{System Description}
The U-net \cite{RonnebergerFB15-unet} model has been selected as the prediction core to perform the trajectory prediction through an image-to-image regression process. This model is similar to a CNN, which was developed to perform semantic segmentation in biomedical imagery. U-Net architecture is based on four different blocks: pre-processing, encoder, decoder, and post-processing. Finally, application layers adapt the network's output to the desired topology problem, such as semantic segmentation or image regression.

The U-net receive as input an image with dimensions $H\times W\times D$, where $H$ is the height, $W$ is the width, and $D$ is the number of channels. Note that BEV representations are single-channel images and consecutive representations are stacked over the channel dimension $D$. 
The pre-processing block generates $K$ features from the input data, producing an output volume with dimensions $H\times W \times K$.
The encoder concentrates spatially the feature volume by reducing the input size by four and increasing the depth by two. The first encoder produces an output size $H/2 \times W/2 \times 2K$. The decoder blocks perform the opposite action, increasing the output size by four and reducing the depth by two. The last decoder produces an output size $H \times W \times K$.
Encoder and decoder blocks are in pairs and the same number of each one is needed to preserve the consistency of the data sizes. The depth of the U-net is defined by the number of encoder-decoder pairs and defined by $n$.
The feature volume generated by an encoder is forwarded to the input of the next encoder and its decoder pair, where it is combined with the output of the previous decoder to combine low-level with high-level features.
The post-processing block is equivalent to the pre-processing block, it generates $M$ output channels from its input volume.

\begin{figure}[ht]
  \centering
  \includegraphics[width=0.8\linewidth]{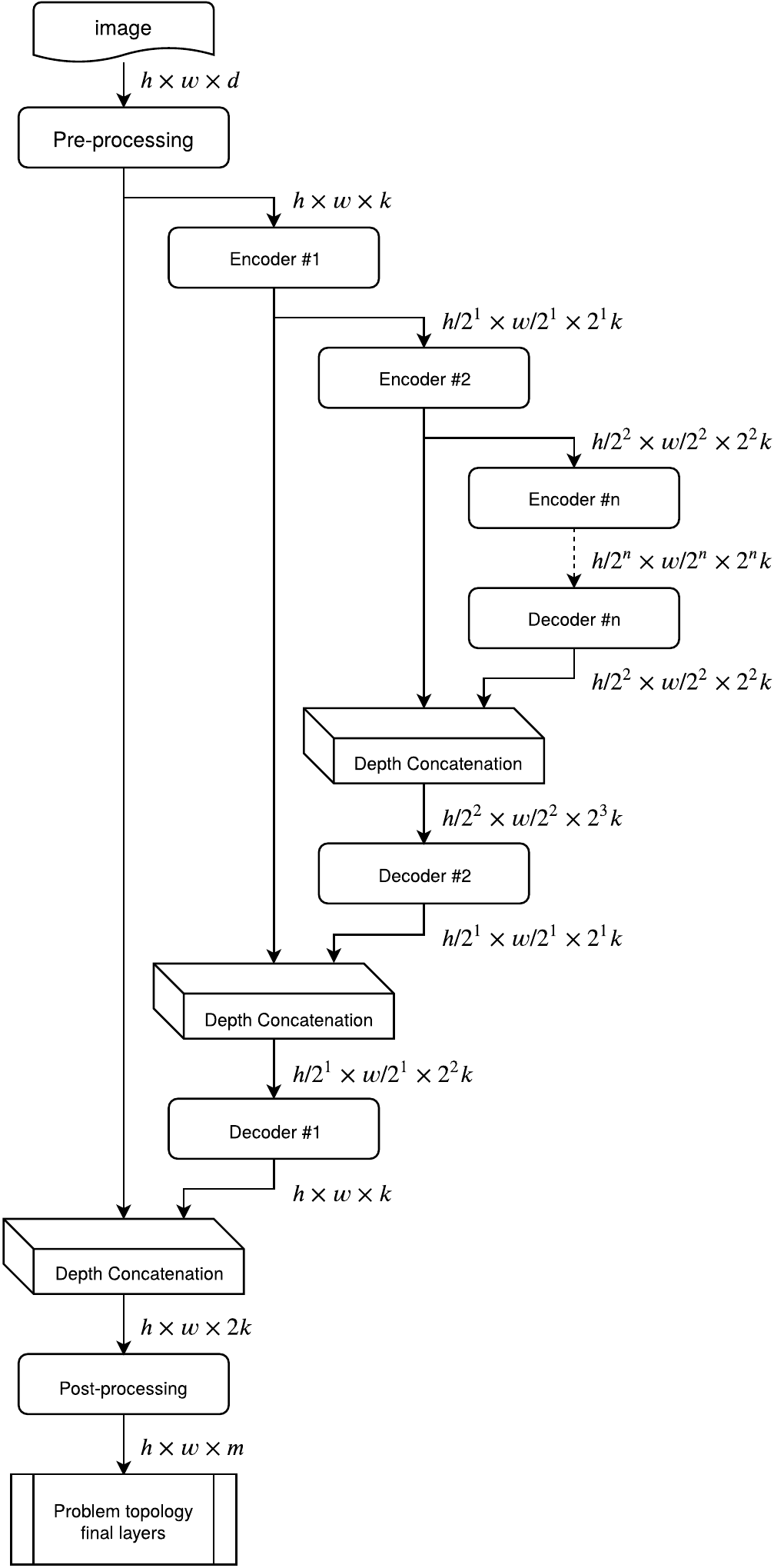}
  \caption{U-net architecture with $n$ depth levels.}
  \label{fig:u-net_architecture}
\end{figure}

Figure \ref{fig:u-net_architecture} shows a generic representation of the U-net architecture, indicating how data sizes are modified by different blocks through $n$ depth levels. It can be seen that the original input image with dimension $H \times W \times D$ is transformed to an $H \times W \times M$ image, where spatial dimensions are preserved and the channels (used as temporal dimension) can be adjusted to set up the desired number of prediction steps.

This network has the particularity of connecting each pixel at the output with its counterpart at the input and an adjacent region determined by the number of depth levels $n$. With an adequate number of depth levels, any pixel at the output can be connected to all pixels at the input, ensuring a total interconnection. In general, the receptive field can be expressed as in \ref{eq:inputreceptive} where $r$ is the semi-length of the contact square. The network architecture and the depth levels $n$ determine the minimum input size that can be expressed as $2^n \times 2^n$. Table \ref{table:unetstruct} summarizes the main features of the U-net model for common depth levels.

\begin{equation}
\label{eq:inputreceptive}
r = \pm2 \left( 3+\sum_{i=2}^{n}{5\cdot 2^{i-2}}\right)
\end{equation}

The receptive field has a crucial role to play, the longer the receptive field the longer the interaction range between vehicles.

\begin{table}[!ht]
\renewcommand{\arraystretch}{1.1}
    \centering
    \caption{U-net structure based on Depth levels. Contact Area, minimum input size, and the number of parameters.}
    \label{table:unetstruct}
    \begin{tabular}{ l | c | c | c }
    Depth levels $n$        & 4         & 5         & 6\\
    \hline
    Contact Area        & $152\times 152$   & $312\times 312$ & $624\times 624$ \\
    Min. input size  & $16\times16$ & $32\times32$ & $64\times64$ \\
    Parameters          & 56k       & 116k      & 235k  \\
    \end{tabular}
\end{table}

\subsection{Input and Output Representation}
Vehicle positions and road configuration are represented in a schematic BEV representation. As mentioned before, BEV representations are single-channel images with dimensions $H \times W$. The transformation between world or local coordinates to an BEV image defines a Pixel per Meter (PPM) conversion factor. This parameter can be equal or different for $X$ and $Y$ dimensions and affects the interaction area between vehicles directly as a result of the conversion between meters and pixels.

The single-channel BEV representations need to manage efficiently the gray-scale values to create understandable images. The vehicles are represented using one of the following options:  
\begin{itemize}
  \item Rectangular representation. It uses a fixed size rectangle $w_v \times h_v$ to populate the BEV with a fixed value $I_v$. This representation is closer to reality, as it matches better vehicles' shape.
  \item Bi-dimensional Gaussian distribution. The Gaussian distribution represents the probability of being a vehicle using a specific tile in the BEV, according to \eqref{eq:gaussian}. 
  The mean value of each bi-dimensional Gaussian distribution $\mathbf{\mu}$ is set using the vehicle's position. The standard deviation $\mathbf{\sigma}$ is composed of half of the rectangle size.
\end{itemize}

\begin{equation}
    I_v(x,y) = I_v\exp{-\left(\left(\frac{x-\mu_{x_i}}{\sqrt{2}\sigma_{x_i}}\right)^2+\left(\frac{y-\mu_{y_i}}{\sqrt{2}\sigma_{y_i}}\right)^2\right)}
    \label{eq:gaussian}
\end{equation}

At the point where two vehicles' representations overlap, there are two options to merge the area shared by them. They can be added, generating values up to $2I_v$, or they can be limited to the maximum of each vehicle representation according to \ref{eq:probmax}. The second option to combine the shared areas represents the real scene in a more reliable way, and the maximum value of the representation keeps limited to $I_v$.

\begin{equation}
  I_v(x,y) = \max_{\forall}{\{ I_{v,i}(x,y)\}}
  \label{eq:probmax}
\end{equation}

Additionally, road structure is represented into the BEV to provide context information. Road markings are represented with a fixed value $I_L$. If the complete image span (from 0 to 255) is used for representation, as in the case of the Gaussian distributions, $I_L$ can match with some points of the vehicle's representation.

\begin{figure}[ht!]
    \includegraphics[width=\linewidth]{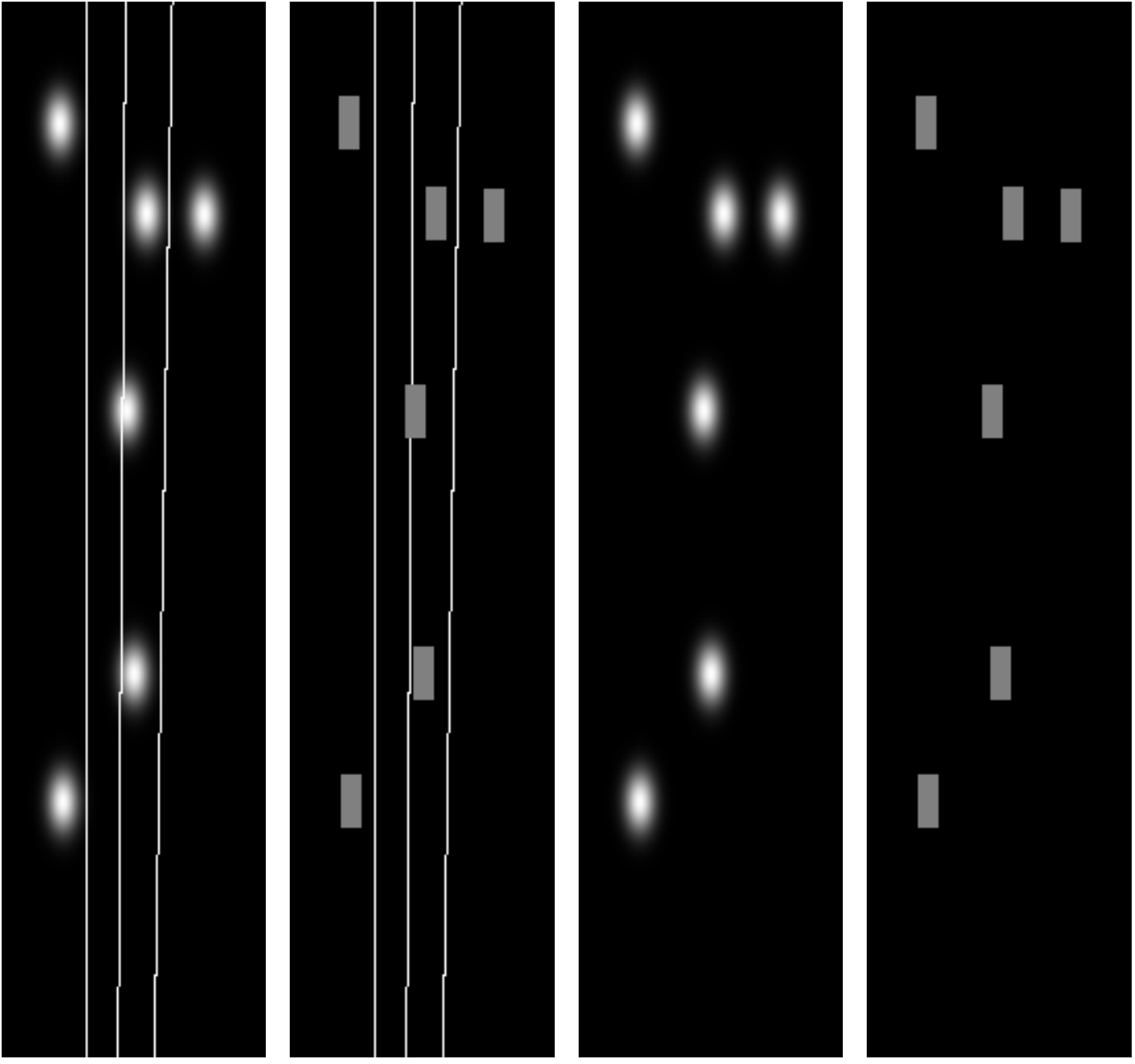}
    \caption{Example of four BEV vehicles' position representation using Gaussian and Rectangular shapes with and without lane information. $H=512, W=256, PPM_x = 0.2, PPM_y=0.1, I_v=255, I_L=255$}
    \label{fig:bev_repre}
\end{figure}

Figure \ref{fig:bev_repre} illustrates an example of the BEV representation. Different combinations of Gaussian and rectangle vehicle representations are combined with lane markings. For this representation, $I_v$ is set to 255 for Gaussian representation and 128 for the rectangular representation. $I_L$ is set to 255 for lane markings representation. Note that the two right images do not show lane markings because of $I_L = 0$. This has been carried out on purpose to illustrate differences between input (two left) and output (two right), as output representations are used to forecast vehicle positions and not lane positions.

Both input and output images are generated in the same way, but output images do not include lane representations. The complete input and output data consist of $D$ consecutive samples stacked by creating an input/output volume with size $H \times W \times D$. When $D$ time samples of data are stacked, a new problem arises in the output block. The output block represents future vehicle positions and three kinds of vehicles coexist:

\begin{itemize}
  \item Vehicles that are present in the input and output block. This is the most common case. In this case, predicting their positions is achievable.
  \item Vehicles that appear in the input block but do not appear in the output block. These are vehicles that abandoned the represented area, thus predicting their positions is unnecessary.
  \item Vehicles that do not exist in the input block, but they do exist in the output block. These are vehicles that enter the study area, therefore the prediction of their positions is unfeasible.
\end{itemize}

Predict positions of vehicles whose existence is unknown is like \textit{performing some magic}. Output representations are generated considering only vehicles present in the latest of the input representations.

\subsection{Vehicle Position Extraction}\label{subsec:extractionalgo}
The codification procedure transforms numeric positions into visual representations. The inverse procedure, transforming visual representations into numeric positions is needed to compute the performance of the prediction process. For each input image, $n$ different predictions are generated at different time horizons. The number of existing vehicles in a future scene representation is \textit{a priori} unknown, so the way used to extract numeric positions must be able to produce an unknown amount of them. It can only consider the vehicles in the future should be the same or fewer than in the latest input representation.

The algorithm proposed in \ref{alg:positionextracion} is used to extract the position of the vehicles. The output data represents the probability of each pixel to be part of a vehicle in a certain future sample. The algorithm finds the pixel with the highest probability first. This pixel is denoted as $P=(R, C)$, and it is used as the discrete location of the vehicle. The proposed algorithm extracts the position with sub-pixel resolution in a second step. The position of the vehicle is refined using a scoring function. Each pixel $p_i$ included in the area defined by a rectangle with dimension $R = 2W \times 2H$ around $P$ contributes weighting the probability by its pixel coordinates according to \eqref{eq:scoringfcn}. Note that discrete positions are conditioned by the resolution used to define the BEV representation.

\begin{equation}\label{eq:scoringfcn}
    \hat P(\hat{R},\hat{C}) = \sum_{r = R-h}^{r = R+h}{\sum_{c = C-w}^{c = C+w}{p(r,c)\cdot(r,c) }}\\
\end{equation}

\begin{figure}[ht!]
    \centering
    \includegraphics[width=\linewidth]{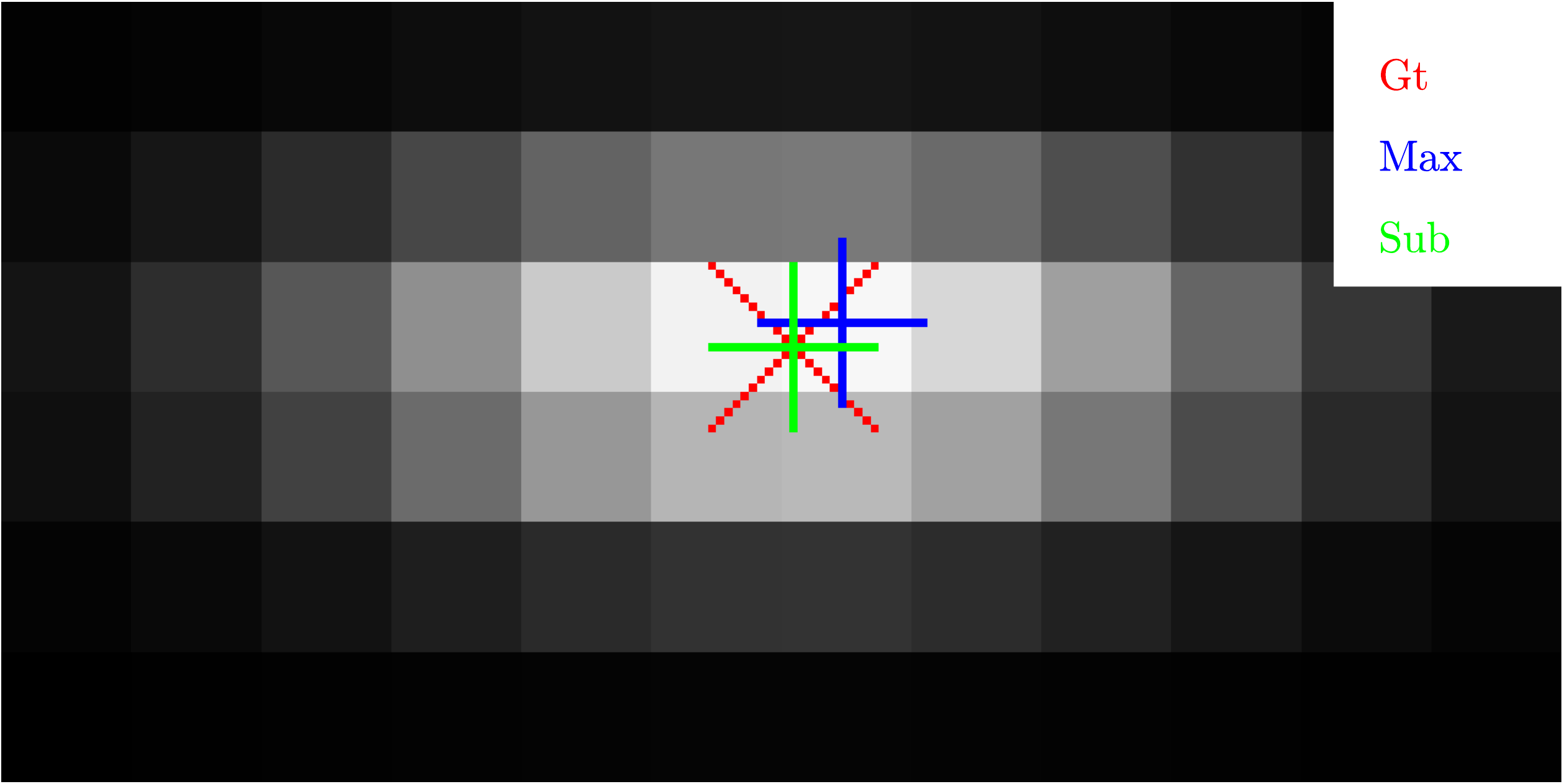}
    \caption{Position extraction from graphic representation. The red cross represents the actual center of the vehicle, the blue plus symbol the discrete found position, and the green plus symbol the sub-pixel position. Image augmented 16 times.}
    \label{fig:subpixel}
\end{figure}

After computing the sub-pixel position, the area used to compute it is cleared, setting the probability to zero. This procedure is repeated as many times as pixels with a probability higher than $p_{min}$ persist in the representation. According to the limits of the representation, being 0 a non-used pixel and 255 the maximum value for a used pixel, $p_{min}$ is set to 128, which is the limit to consider that a pixel represents a possible vehicle. Figure \ref{fig:subpixel} shows the codification of an arbitrary vehicle using the Gaussian representation. The red cross represents the actual center of the vehicle, the blue plus symbol represents the discrete found position of the vehicle, and the green one is the obtained sub-pixel position. Note that the image has been amplified by 16 to illustrate the differences between discrete and sub-pixel detection.

\begin{algorithm}
\small
   \caption{Multi-Vehicle Location Extraction}\label{alg:positionextracion}
    \begin{algorithmic}[1]
      \Function{$\mathcal P =$ Extraction}{$p(r,c),p_{min},w,h$}
        \State $\mathcal{P} = \emptyset$
        \While{$\exists(r,c) \mid p(r,c) > p_{min}$}
            \State $P = (R,C) \mid p(R,C) > p(r,c) \forall (r,c)$
            \State $\hat{P}(\hat{R} ,\hat{C}) = SubPxLocation(p(r,c), w, h, P)$
            \State $\hat{P} \in \mathcal{P}$
            \State $\forall(r,c)\in P \pm (h,w)$ do $p(r,c) = 0 $
        \EndWhile
      \EndFunction
      
      \Function{$ \hat P =$ SubPxLocation}{$p(r,c), w, h, P$}
      \State $\hat R = \hat C = 0$
      \For{$R-h < r < R + h$} 
        \For{$C-w < c < C + w$}
            \State $\hat R += p(r,c)\cdot r$
            \State $\hat C += p(r,c)\cdot c$
        \EndFor
      \EndFor
      \State $\hat P=( \hat R / 2h, \hat C / 2w)$
      \EndFunction
\end{algorithmic}
\end{algorithm}

Table \ref{table:extraction} shows the position extracted from the vehicle shown in figure \ref{fig:subpixel}. The vehicle is represented in the BEV using the same resolution in both axes, $PPM_x = PPM_y = 1$. The position of the vehicle is at coordinates $X=6.63$ and $Y=3.21$. The resolution used in the representation defines the error generated for the maximum method. However, the sub-pixel resolution method shows an error 100 times lower than the resolution parameter.


\begin{table}[ht]
\renewcommand{\arraystretch}{1.1}
    \centering
    \caption{Error of extracted vehicle position for Maximum and Sub-pixel methods.}
    \label{table:extraction}
    \begin{tabular}{ l | c | c }
                                & X / Y [m]/[m]  & X / Y Error [m]/[m]   \\
        \hline
        Original                & 6.63 / 3.21   & - / -                 \\
        Maximum                 & 7 / 3         & 0.37 / 0.21           \\
        SubPixel                & 6.615 / 3.216 & \textbf{0.015 / 0.006}\\
    \end{tabular}
\end{table}

\subsection{Association of Extracted Positions}\label{subsec:association}
Extracted positions need to be associated with the corresponding vehicle. A simple procedure based on a Hungarian matrix \cite{kuhn1955kuhn} is used to associate the extracted positions with the positions used in the latest input representation. The number of elements that can be matched is the minimum between the number of extracted positions from the output image or the number of vehicles represented in the latest input representation. The value used as the distance parameter to associate elements is the Euclidean distance between extracted points and the provided samples. This method is good enough as predicted positions do not differ from their current positions.

\subsection{Training Strategy} \label{section:training}
The training procedure was carried out using the PREVENTION dataset \cite{izquierdo2019prevention} which includes on-board vehicle detections on highway scenarios. The dataset consists of 11 sequences recorded at 16 Hz. The number of frames accounts for a total of 345k samples representing more than 6 hours of traffic recordings. The amount of data is massive and allows us to train models with a wide variety of samples and situations. 

The original frame rate has been reduced from 16Hz to 4Hz, discarding three consecutive samples out of four because it is too high to appreciate relevant displacements from one sample to the next one. 
The lowered frame rate allows the input and output data to cover a larger period using the same number of samples. The number of consecutive input and output samples has been set to 8 ($D=M=8$). According to the lowered data rate the input block represents the period from t - 1.75 seconds to t and the output block represents future vehicle locations in the period from t + 0.25 to t + 2.0 seconds. The range of the vehicle positions contained in the dataset is limited to 100 meters in longitudinal and up to 12 meters in both lateral directions. The BEV representation has been defined as a grid with $H=512$ by $W=256$ pixels representing an area of 102,4 by $\pm$12,8 meters, respectively. The size of the grid has been established based on three criteria: memory size when it is allocated in the GPU for training purposes, a convenient resolution to understand the scene, and compatibility with the proposed network architecture. The lateral and longitudinal resolution is $PPM_X = 5$ and $PPM_Y = 10$ respectively. The lateral resolution is bigger in order to amplify lateral displacements. Rectangle size $w_v \times h_v$ for vehicle representation has been set to standard vehicle dimensions $1.8\times 5.0$ meters. Equivalently, $\mathbf{\sigma} = (0.9,2.5)$.

\begin{figure*}[h!]
  \centering
  \includegraphics[width=\linewidth]{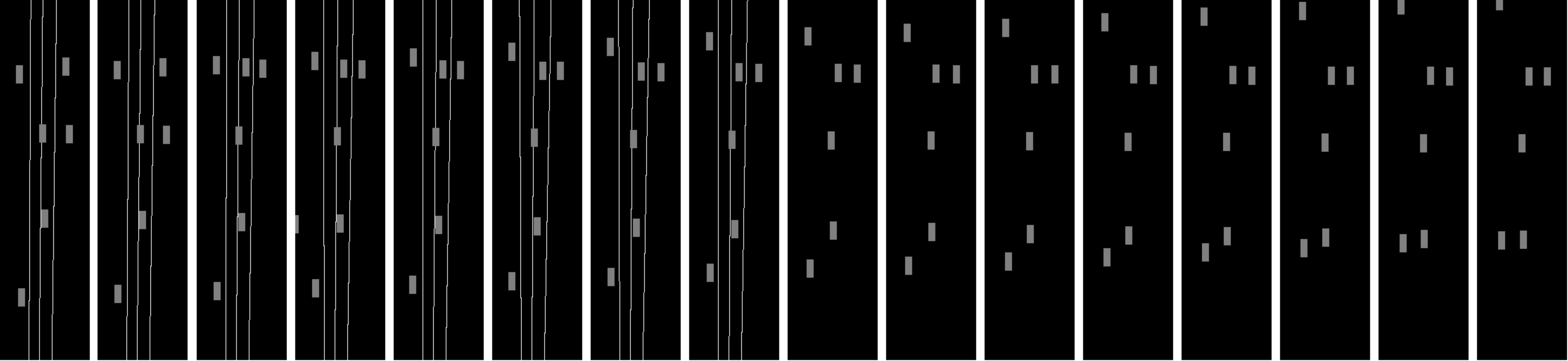}
  \caption{Example of BEV input (8 leftmost) and output (8 rightmost) block sequences. This sequence represents a total of 4.0 seconds of trajectories. From t-1.75 seconds to t as input data and from t to t+2.0 seconds as output data.}
  \label{fig:bev_io_seq}
\end{figure*} 

The dataset was split from sequence 1 to 9 as the training set and from sequence 9 to 11 as the test set. The hyper-parameters used in the training process were: optimizer Adam, mini-batch 1, epochs 1, initial learning rate $10^{-6}$, L2 regularization $10^{-4}$, momentum 0.9, and gradient threshold 1. The Root Mean Square Error (RMSE) has been used as a loss function for the image-to-image regression problem.

Several trainings have been conducted to figure out the best performing configuration. The effect of the rectangle and Gaussian vehicle representations, the use of lane markings, the U-net depth levels, and the terminal layer of the network has been evaluated. Input and output images with discrete values ranging from 0 to 255 have been converted into floating-point values ranging from 0 to 1 for the training process.

The tested U-net's depth levels were limited to 4, 5, and 6. Level 7 and above exceeded the GPU memory size and the training could not be conducted. Depth levels below 4 were discarded, as they offered a very limited contact area.

Three different terminal layers were tested in these experiments. The objective was to identify the best-performing one. The \textit{linear} layer does not apply a transformation to the network's output. The \textit{tanh} layer applies the hyperbolic tangent function ranging the output from -1 to 1, with a nonlinear transformation. The \textit{clippedReLU} layer limits the output below 0 and above a user-defined value, which is set to 1 for this problem.

Figure \ref{fig:bev_io_seq} illustrates a complete input-output sequence. For simpler understanding vehicles have been represented as rectangles using the same $PPM$ factor for both axes. Lanes are only used in the input set of images (8 leftmost images). The output images (8 rightmost) have no lane markings, given that the desired output is the vehicles' positions only.

\subsection{Baseline}\label{subsec:baseline}
For comparison purposes, a KF with a \textit{Constant Speed} model has been used to process vehicle positions and generate predictions.
State vector $X$, and the observation vector $Y$, are defined in \ref{eq:statevector} where $x$, and $y$ are vehicle positions, $v_x$ and $v_y$ are vehicle speeds in both axis. The process model $A$, and the observation model $H$ are described in \ref{eq:kfmodel1} and \ref{eq:models} where $\Delta t$ represents the time step. 

\begin{equation}
  \label{eq:statevector}
  X =
  \begin{bmatrix}
  x &  y &  v_x & v_y 
  \end{bmatrix}^{\mathrm{T}}
  \,\,\,\,\,\,\,\,\,\,
  Y = 
  \begin{bmatrix}
    x & y & v_x & v_y 
  \end{bmatrix}^{\mathrm{T}}
\end{equation}

\begin{equation}
  \label{eq:kfmodel1}
  \begin{split}
  X_k = A X_{k-1} \,\,\,\,\,\,\,\, Y_k = H X_{k}
  \end{split}
\end{equation}
\begin{equation}
  \label{eq:models}
  A = 
  \begin{bmatrix}
    1 & 0 & \Delta t &  0 \\
    0 & 1 &  0 & \Delta t \\
    0 & 0 &  1 &  0 \\
    0 & 0 &  0 &  1 \\
  \end{bmatrix}
  \,\,\,\,\,\,\,\,\,\,
  H =  
  \begin{bmatrix}
    1 & 0 & 0 & 0 \\
    0 & 1 & 0 & 0 \\
    0 & 0 & 1 & 0 \\
    0 & 0 & 0 & 1 \\
  \end{bmatrix}
\end{equation}

\section{Trajectory Prediction Results}\label{sec:trajectory_results}

This section presents the results generated by the proposed trajectory prediction model. This section is divided into two subsections, subsection \ref{subsec:trajectory_metrics} provides quantitative results using different commonly accepted metrics and subsection \ref{subsec:trajectory_cualitative} presents graphic representations as qualitative results.

\subsection{Quantitative Results}\label{subsec:trajectory_metrics}
The quantitative results of trajectory prediction are evaluated in this subsection. For trajectory prediction evaluation, two metrics are used to evaluate the quality of each model. The RMSE and the MAE are used as performance metrics and they are provided in both longitudinal and lateral coordinates.  

The RMSE is computed according to \eqref{eq:rmse} where subindex $k$ denotes each prediction step and subindex $i$ represents each individual trajectory in a set with a total of $N$ trajectories.

\begin{equation}
     \label{eq:rmse}
     RMSE_k = \sqrt{\sum_{i=0}^{N}\dfrac{(\hat{x}_{k,i}-x_{k,i})^2}{N} }
\end{equation}

The Mean Absolute Error (MAE) represents the unweighted average error as it is defined in \eqref{eq:mae}.
\begin{equation}
     \label{eq:mae}
     MAE_k =  \sum_{i=0}^{N} \dfrac{\abs{\hat{x}_{k,i}-x_{k,i}}}{N} 
\end{equation}

The literature evaluates commonly the performance of trajectory predictive models providing the Average Displacement Error (ADE) and the Final Displacement Error (FDE). In this work, RMSE and MAE are provided for each prediction step, consequently, FDE is the MAE at the last prediction step and the ADE is directly computed as the MAE's average value according to \eqref{eq:ate} where $M$ is the total number of prediction steps.

\begin{equation}
     \label{eq:ate}
     ADE = \dfrac{1}{M}\sum_{k=1}^{M} {MAE_k}
\end{equation}

The input block provides a video sequence representing 1.75 seconds of past information and the predictive model generates an output block that represents the scene 2.0 seconds in advance. This output block has 8 samples at 0.25 second time intervals. 

First, all the configurations related to the U-net architecture were tested. For these trainings, the Gaussian representation of vehicles with no lane markings was used. Table \ref{table:traj_pred_rmse} presents the RMSE for these configurations at different prediction intervals. U-net has been tested with 4, 5, and 6 depth levels together with the \textit{linear} and \textit{clippedReLU} final layers. The \textit{tanh} configuration produced unstable trainings, generating divergence and aborting because of computation errors. This configuration has been removed from tables due to their consistent null results. U-net models are denoted as U-net-4 for the configuration with 4 depth levels, and similarly for 5 and 6 depth levels.
Longitudinal and lateral error are denoted as $\varepsilon_x$ and $\varepsilon_y$, respectively and expressed in meters for all tables.

First row in table \ref{table:traj_pred_rmse} presents results for KF which is used as baseline method for comparison purposes. The KF has been used to predict positions at the same prediction horizons as the CNN-based models.

\setlength{\tabcolsep}{2pt}
\begin{table}[ht]
\footnotesize
\renewcommand{\arraystretch}{1.1}
\centering

\caption{Trajectory prediction error as RMSE for different network depth and terminal layer configurations.}
\label{table:traj_pred_rmse}
     \begin{tabular}{ l | c | c | c }
                                                & $t = 0.25$                    & $t = 1.0$                     & $t = 2.0$                    \\
     Model                                 & $\varepsilon_x\;\;/\;\;\varepsilon_y$ & $\varepsilon_x\;\;/\;\;\varepsilon_y$ & $\varepsilon_x\;\;/\;\;\varepsilon_y$\\
     \hline 
     KF                                    & 1.20 / 0.47                   & 1.76 / 0.94                   & 1.98 / 1.44                  \\
     U-net-4, \textit{linear}          & 1.24 / 0.65                   & 1.65 / 0.97                   & 2.39 / 1.40                  \\
     U-net-4, \textit{clipReLU}     & 1.36 / 0.71                   & 1.68 / 1.04                   & 2.51 / 1.39                  \\
     U-net-5, \textit{linear}          & 0.43 / 0.23                   & 0.62 / 0.55                   & 1.06 / 0.81                  \\
     U-net-5, \textit{clipReLU}     & 0.74 / 0.38                   & 0.95 / 0.68                   & 1.93 / 0.94                  \\
     \textbf{U-net-6, \emph{linear}} & \textbf{0.35 / 0.22}          & \textbf{0.56 / 0.52}          & \textbf{0.93 / 0.69}         \\
     U-net-6, \textit{clipReLU}     & 0.65 / 0.27                   & 0.94 / 0.71                   & 1.72 / 0.87                  \\
     \end{tabular}
\end{table} 
\setlength{\tabcolsep}{6pt}

Comparing the effect of the terminal layers there is a huge difference between experiments. The error is nearly half for the \textit{linear} layer compared with the \textit{clippedReLU}. This difference gets smaller when less complex models where fewer depth levels are used. A priori, the \textit{clippedReLU} was expected to offer a better fitting for the output data. However, this layer introduces a zero gradient area out of the defined response interval (0-1).

\begin{figure}[ht]
     \centering
      \includegraphics[width=\linewidth]{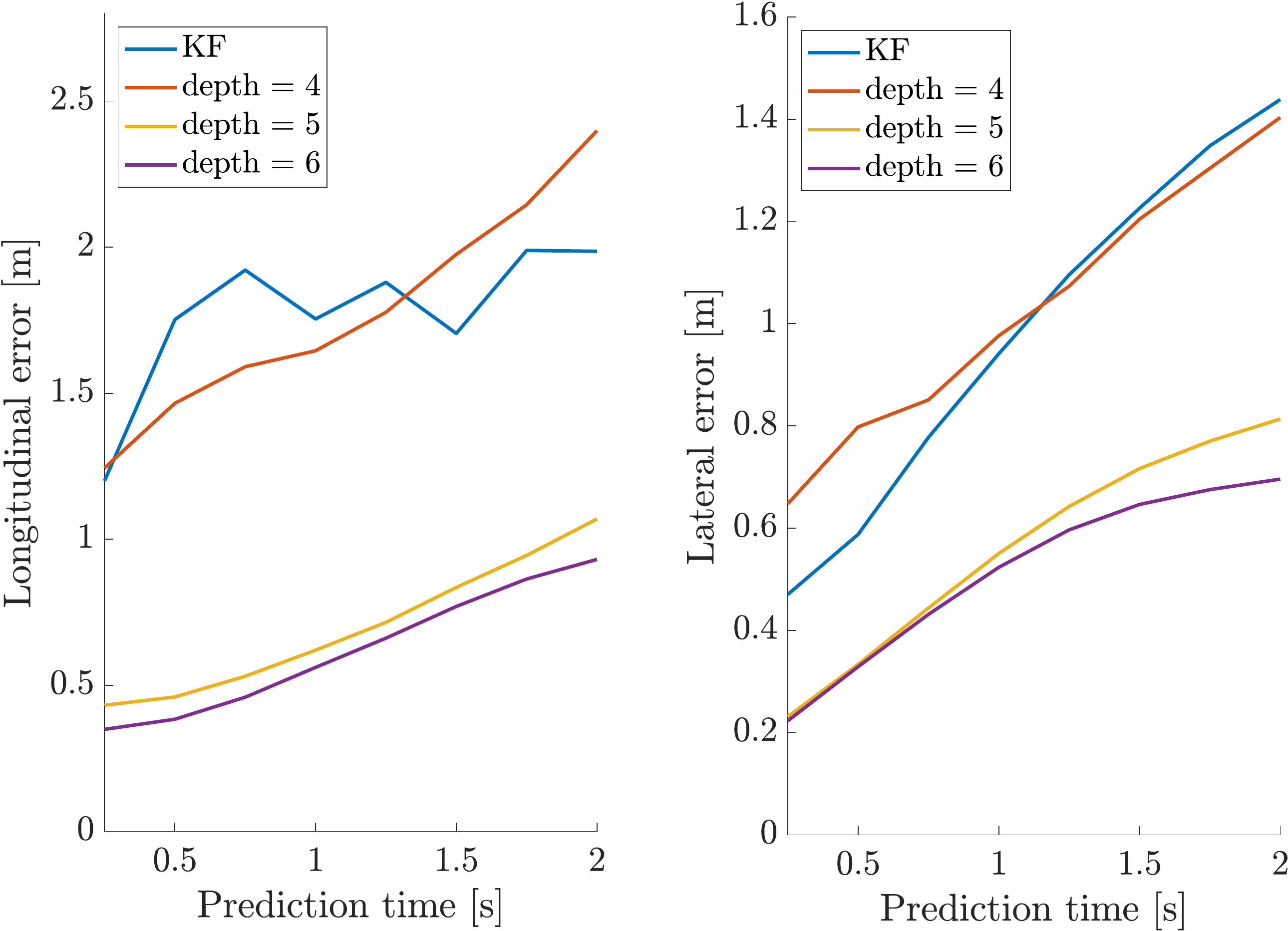}
      \caption{Effect of depth levels in the trajectory prediction performance. Longitudinal (left) and lateral (right) RMSE for different depth levels using the \textit{linear} terminal layer.}
      \label{fig:rmse}
\end{figure}

Figure \ref{fig:rmse} shows the RMSE for the baseline method and the U-net model with the \textit{linear} terminal layer and different depth levels as a visual complement of the values presented in table \ref{table:traj_pred_rmse}. It can be observed that the deeper the model the lower the RMSE. The best performing model is the U-net-6 with the \textit{linear} terminal layer, followed closely by the U-net-5 with the same configuration. U-net-4 and the baseline method present similar results.

The similarity of the U-net-4 and the KF performance can be explained because the U-net-4 has a small receptive field and the future positions of the vehicles are inferred based only on near objects, basically the ego vehicle itself. U-net-5 or U-net-6 increases their receptive field exponentially enabling vehicle interactions. This could be the reason why they perform better than the KF, which does not include any other information than vehicle dynamics.

U-net-6 prediction errors are 53\% and 52\% lower compared to the baseline method in longitudinal and lateral coordinates for 2.0 seconds prediction horizon. The U-net-5 prediction errors are 46\% and 44\% lower for the longitudinal and lateral errors respectively.

Table \ref{table:traj_pred_mae} shows the MAE for the \textit{linear} set of U-net configurations only. As an unweighted metric, it is easier to understand expected errors at each prediction step. Moreover, ADE and FDE are provided as common literature metrics.

\setlength{\tabcolsep}{2pt}
\begin{table}[ht]
\footnotesize
\renewcommand{\arraystretch}{1.1}
\centering
\caption{Trajectory prediction error as a Mean Absolute Error for different network depth configurations using the linear terminal layer.}
\label{table:traj_pred_mae}
     \begin{tabular}{ l | c | c | c | c}
                                             &                               &                               & FDE                          & ADE  \\
                                             & $t = 0.25$                    & $t = 1.0$                     & $t = 2.0$                     &    \\
     Model                        & $\varepsilon_x\;\;/\;\;\varepsilon_y$ & $\varepsilon_x\;\;/\;\;\varepsilon_y$ & $\varepsilon_x\;\;/\;\;\varepsilon_y$ & $\varepsilon_x\;\;/\;\;\varepsilon_y$\\
     \hline 
     KF                                      & 0.24 / 0.22                   & 0.58 / 0.65                   & 1.13 / 1.04                    & 0.67/0.68 \\
     U-net-4 & 0.35 / 0.18                   & 0.57 / 0.46                   & 1.43 / 0.85                    & 0.69/0.49   \\
     U-net-5 & 0.23 / 0.15                   & 0.44 / 0.39                   & 0.84 / 0.59                    & 0.51/0.40    \\
     \textbf{U-net-6}  & \textbf{0.20 / 0.14}          & \textbf{0.42 / 0.38}          & \textbf{0.76 / 0.53}           & \textbf{0.47/0.38}\\
     \end{tabular}
\end{table} 
\setlength{\tabcolsep}{6pt}

MAE values have been notably reduced compared with the RMSE values. However, the RMSE arguments are valid for the MAE. The deeper the network the better the predictions. KF's errors are comparable to the U-net-4 model again. As remarkable points, the ADE for the U-net-6 is 0.51 and 0.40 meters for lateral and longitudinal respectively. The FDE for the U-net-6 reaches 0.76 and 0.53 meters in longitudinal and lateral errors on average for a 2.0 seconds prediction horizon.

The MAE of U-net-6 with the \textit{linear} terminal layer is 33\% and 49\% lower in longitudinal and lateral with respect to the baseline method for 2.0 seconds prediction horizon. If the ADE is used to compare models, the U-net-6's ADE is 30\% and 44\% lower than the KF's ADE for longitudinal and lateral coordinates respectively.

After the evaluation of different U-net configurations, the best performing model (U-net-6 and \textit{linear} terminal layer) is tested with different scene representations in order to figure out the best one. Vehicle Gaussian and rectangular representations together with or without lane markings have been evaluated. Table \ref{table:traj_pred_input_mae} shows the MAE, FDE, and ADE for these configurations.

\setlength{\tabcolsep}{2pt}
\begin{table}[ht]
\footnotesize
 \renewcommand{\arraystretch}{1.1}
 \centering
 \caption{Trajectory prediction error as a Mean Absolute Error for different input configurations using the U-net-6 model with the linear terminal layer.}
 \label{table:traj_pred_input_mae}
      \begin{tabular}{ l | c | c | c | c}
                        &               &               & FDE           & ADE  \\
                          & $t = 0.25$                    & $t = 1.0$                     & $t = 2.0$                     &    \\
      Input Config.            & $\varepsilon_x\;\;/\;\;\varepsilon_y$ & $\varepsilon_x\;\;/\;\;\varepsilon_y$ & $\varepsilon_x\;\;/\;\;\varepsilon_y$ & $\varepsilon_x\;\;/\;\;\varepsilon_y$\\
      \hline
      Gaussian          & 0.20 / 0.14   & 0.42 / 0.38   & 0.76 / 0.53   & 0.47/0.38\\
      Rectangle         & 0.26 / 0.14   & 0.48 / 0.39   & 1.04 / 0.59   & 0.53/0.41\\
      Gauss.+lanes      & 0.19 / 0.16   & 0.43 / 0.40   & 0.75 / 0.56   & 0.48/0.41\\
      Rect.+lanes       & 0.24 / 0.15   & 0.49 / 0.37   & 1.07 / 0.55   & 0.53/0.40\\
      \end{tabular}
\end{table} 
\setlength{\tabcolsep}{6pt}

\begin{figure*}[h!]
  \centering
  \includegraphics[width=\linewidth]{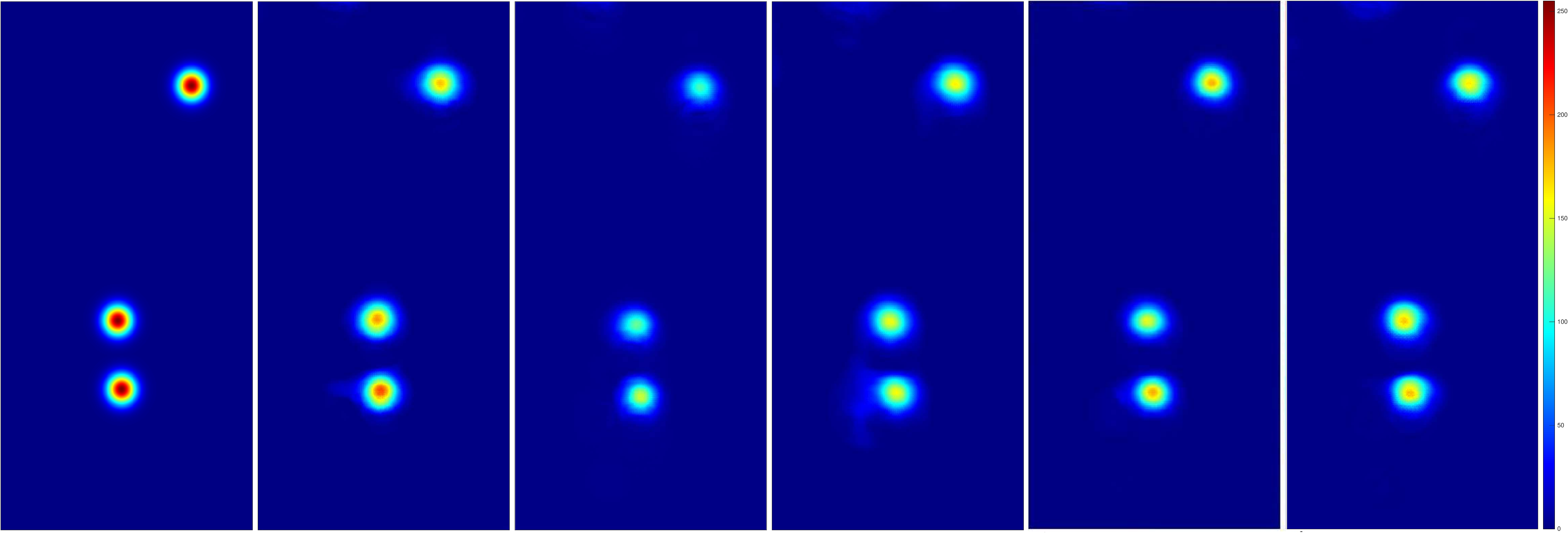}
  \caption{Network configuration and Input representation's effect on output images. From left to right: target output, the \textit{Linear} layer with the Gaussian representation, the \textit{ClippedReLU} layer with the Gaussian representation, the \textit{Linear} layer with the rectangle representation, the \textit{Linear} layer with the Gaussian and lanes representation, and the \textit{Linear} layer with Rectangles and lanes representation.}
  \label{fig:linear_vs_clipped}
\end{figure*}


As observed, the rectangular representation does not provide any improvement compared with the Gaussian representation of vehicles. This fact can be explained because the Gaussian representation models the probability of being a vehicle using a specific area, which seems more robust, in contrast with the binary input of the rectangular representation.

The representation of lane markings in the input images does not produce any significant effect on the prediction performance. It was expected that its use would reduce the prediction error, but no significant changes have been observed. Lane markings are represented by single-pixel lines to avoid occluding vehicle positions. This fact may be the cause to produce no performance changes.

In conclusion, the U-net model trained with 6 depth levels and the \textit{linear} terminal layer using the vehicle's Gaussian representation produces the lowest errors from all the trained models. This configuration overcomes in both longitudinal and lateral coordinates the KF model which has been used as a baseline. U-net with 7 or more depth levels could produce better results due to the observed trend, however, it is impossible to be trained with the currently available hardware.

\subsection{Qualitative Results}\label{subsec:trajectory_cualitative}
This subsection provides qualitative results and examples of the scenes predicted using the U-net architecture.

Figure \ref{fig:linear_vs_clipped} shows predictions 2.0 seconds ahead using different network configurations and input representations. In general, the higher the peak value (warmer color) and rounded the output objects the better the predictions.

Figure \ref{fig:linear_vs_clipped}-I shows the desired output for this particular prediction example. The positions of each vehicle are denoted by a 255 intensity peak value. This is the output that the predictive model must generate at a certain time step based on the corresponding input. The remaining images show the prediction for a U-net-6 with different input and terminal layers. Figure \ref{fig:linear_vs_clipped}-II shows the output generated with the \textit{linear} terminal layer and the Gaussian vehicle representation in the input block. This is the most likely desired output compared with the remaining representations, and numerically it is, as it was exposed in tables \ref{table:traj_pred_input_mae} and \ref{table:traj_pred_mae}. The following representations, from figure \ref{fig:linear_vs_clipped}-III to \ref{fig:linear_vs_clipped}-VI, show configurations with worse results both numerically and visually. Figure \ref{fig:linear_vs_clipped}-III shows the effect of the \textit{clippedReLU} terminal layer. The prediction seems quite similar for both terminal layers, however, the image loses definition, and positions extracted with the \textit{linear} layer are two times more precise. Figure \ref{fig:linear_vs_clipped}-IV shows differences between the Gaussian and the rectangle representation of vehicles. The output image has lower definition but still performs quite well in comparison with the \textit{clippedReLU} variation. Figure \ref{fig:linear_vs_clipped}-V shows the output including lanes together with the Gaussian representation. The definition is similar whether or not the lanes are used. Finally, figure \ref{fig:linear_vs_clipped}-VI shows the output generated using rectangles and lanes in the input block. As when using Gaussian representation, the existence of lanes does not provide any improvement.

\begin{figure}[ht]
     \centering
      \includegraphics[width=\linewidth]{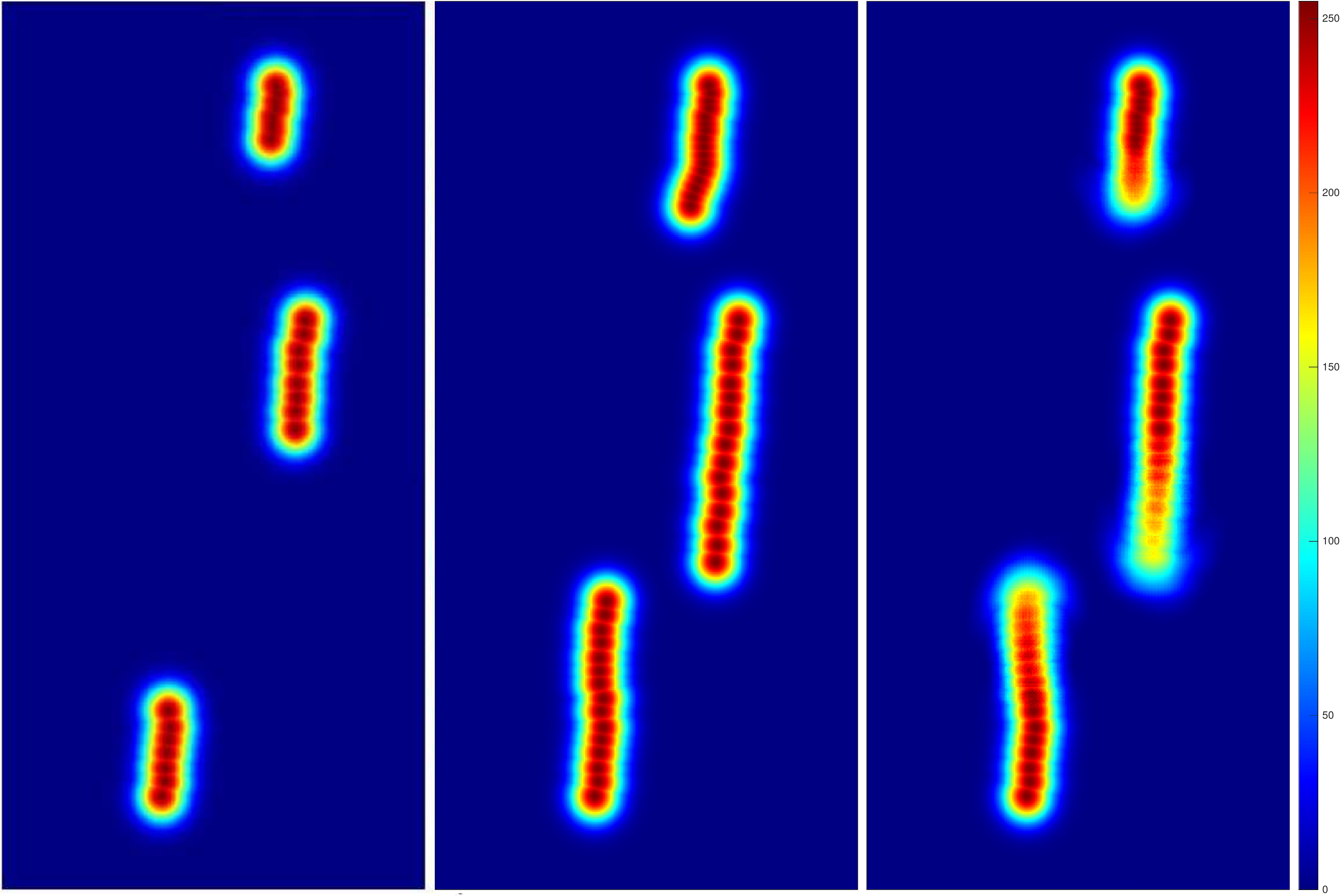}
      \caption{Time series visualization of a trajectory example. From left to right representation of the input positions, the input and the desired positions, and finally, the input and the predicted positions.}
\label{fig:worms}
\end{figure}

Figure \ref{fig:worms} shows an example of a sequence prediction. The images shown in this figure are cumulative past and future samples on a heat map representation. The trajectory of each vehicle is represented as a kind of worm where the first half (8 dots) represents past positions and the second part (8 dots) the future positions. Left image shows the first part of the trajectory which corresponds to the input data used to predict the trajectories. Central image shows the input data representation together with the expected output data, this is the ground truth. Right image shows the input and the predicted positions. Predictions become less defined for longer prediction periods.

Note that this particular sequence shows three vehicles driving at different speeds compared with the ego-vehicle, otherwise, consecutive vehicle positions would be stacked in a small area and trajectories cannot be appreciated.

\begin{figure}[ht]
     \centering
      \includegraphics[width=\linewidth]{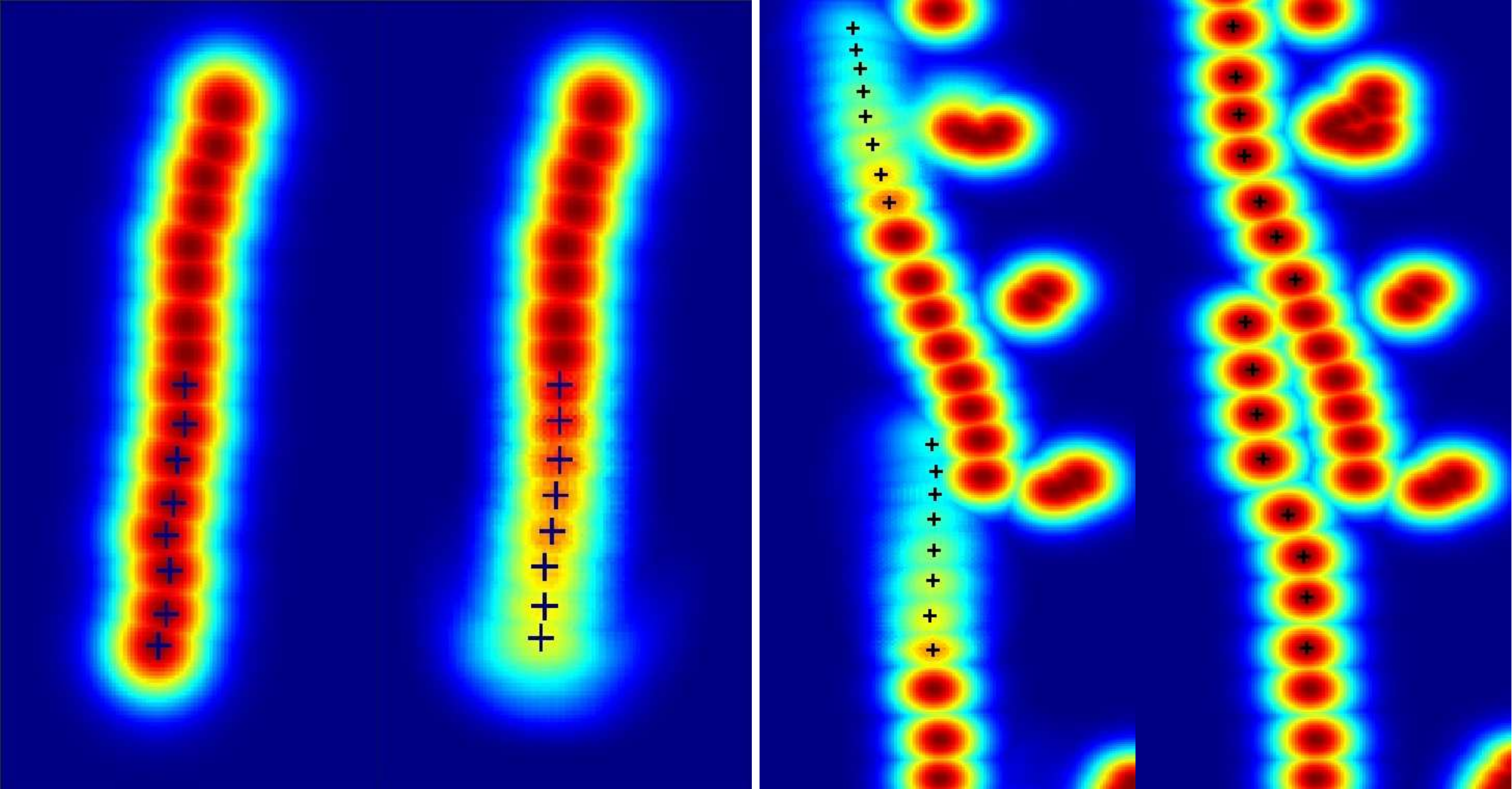}
      \caption{Detail of trajectory prediction. From left to right images 1 and 3 represents the input data and the desired output denoted by $+$ symbols for two different trajectories. Images 2 and 4 represents the same input data and the predicted positions denoted by $+$ symbols for the same trajectories showed in images 1 and 3 respectively.}
\label{fig:worms2}
\end{figure}

Figure \ref{fig:worms2} shows a detailed view of two trajectories, one corresponding with the sequence above and the other is an example that illustrates the level of generalization for predicting lane changes. The two left images are a long trail sequence in which the FDE is 0.22 and 0.24 meters for longitudinal and lateral directions respectively. The predictions become less defined with higher prediction horizons. The two images on the right show two lane changes with a large displacement. Note that the zoom level is lower for these images. It can be seen that the predicted trajectory follows a natural path that avoids obstacles and tends to stabilize in a direction parallel to the direction of movement.

\begin{figure}[ht!]
    \centering
    \includegraphics[width=\linewidth]{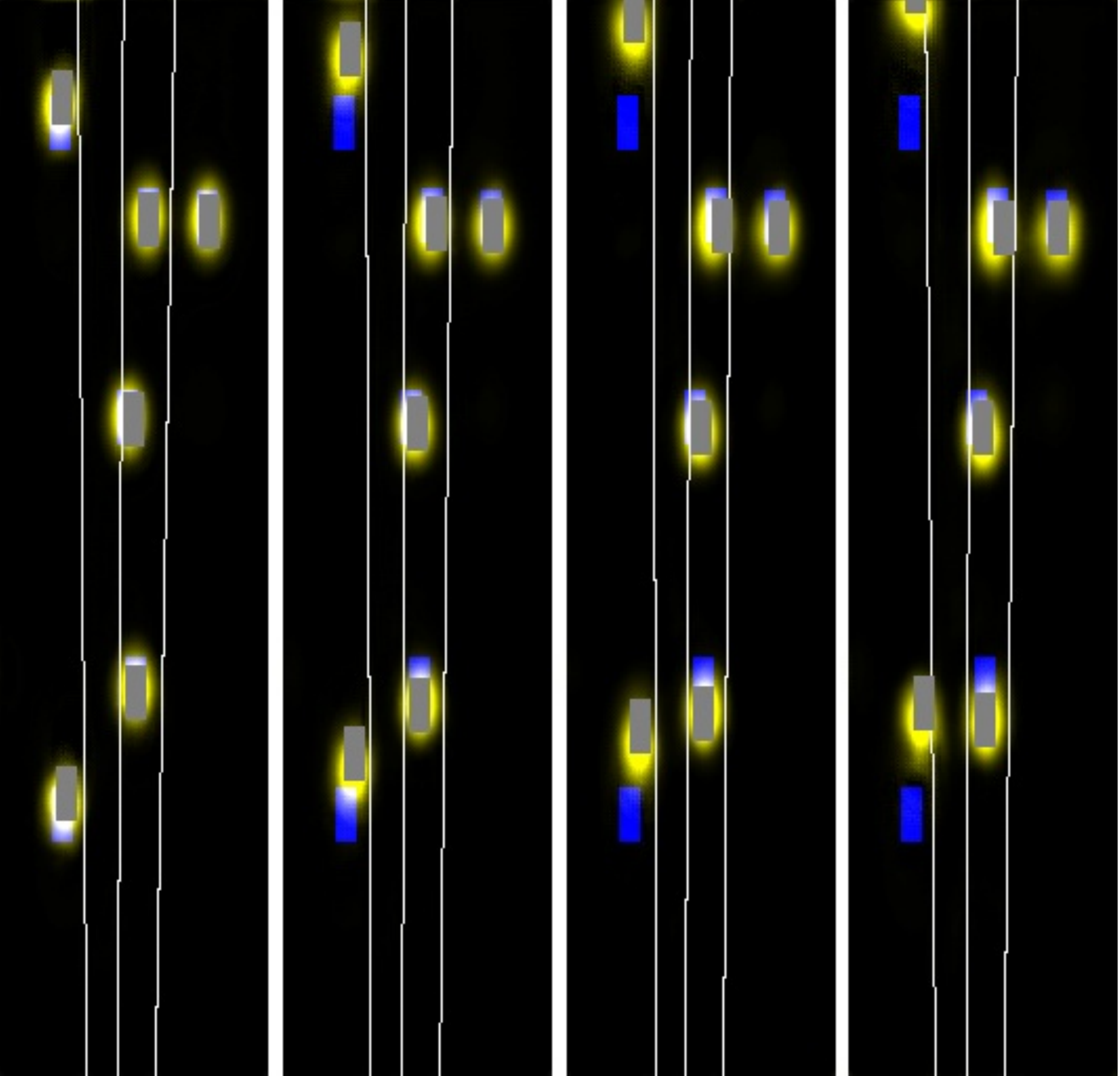}
    \caption{Trajectory prediction results for different prediction horizons (from left to right t + 0.25, t + 1.0, t + 1.5, and t + 2.0 seconds). Last known vehicle positions represented with blue rectangles, current vehicle positions represented with gray rectangles, and predicted vehicle positions represented by yellow blobs. Predictions were performed with the U-net-6 with the \textit{linear} terminal layer and using the Gaussian vehicle representation in the input representation.}
    \label{fig:trajectory_final}
\end{figure}

Finally, figure \ref{fig:trajectory_final} shows a sort of samples in a sequence, using both input and output representation to understand the scene as better as possible. Note that lanes have been used in the predictions only for representation purposes. The actual position of the vehicle is represented with a gray rectangle in all images. The last known position of each vehicle is represented with a blue rectangle. Predicted positions of vehicles are denoted by the yellow Gaussian distributions. The center of each vehicle prediction corresponds with the mass-center of each Gaussian according to algorithm \ref{alg:positionextracion}. Images have been converted back to their natural aspect ratio.

\subsection{Benchmarking}
The proposed model has been trained using the HighD dataset \cite{highDdataset}, which features top-view static-sensor recorded scenes in German highways. This dataset has become popular among trajectory prediction works and is commonly used as a benchmark. For comparison purposes, the input and output data has been subsampled from 25 Hz to 5 Hz, covering a 3 seconds prediction horizon. Table \ref{table:results_highd} presents the RMSE for both the longitudinal and lateral coordinates using the HighD dataset for different depth levels and terminal layer configurations. 

\setlength{\tabcolsep}{2pt}
\begin{table}[ht]
\footnotesize
     \renewcommand{\arraystretch}{1.1}
         \centering
         \caption{Trajectory prediction error as Root Mean Square Error for different network depth and terminal layer configuration using the HighD dataset.}
         \label{table:results_highd}
         \begin{tabular}{ l | c | c | c }
                                               & $t = 1.0$                 & $t = 2.0$             & $t = 3.0$  \\
             Predictive model                  & $\varepsilon_x\;\;/\;\;\varepsilon_y$ & $\varepsilon_x\;\;/\;\;\varepsilon_y$ & $\varepsilon_x\;\;/\;\;\varepsilon_y$\\
             \hline
             U-net-5, \textit{linear}          & 1.17 / 0.27               & 1.57 / 0.36           & 2.36 / 0.54\\
             \textbf{U-net-6, \textit{linear}} & \textbf{0.53 / 0.02}      & \textbf{0.82 / 0.04}  & \textbf{1.23 / 0.07}\\
             U-net-5, \textit{clipReLU}     & 1.35 / 0.52               & 1.67 / 0.63           & 2.51 / 0.94\\
             U-net-6, \textit{clipReLU}     & 1.01 / 0.33               & 1.37 / 0.41           & 2.06 / 0.62\\
         \end{tabular}
\end{table} 
\setlength{\tabcolsep}{6pt}

Note that the results are in the same line as those achieved with the PREVENTION dataset (see table \ref{table:traj_pred_rmse}). The best performing configuration is the U-net-6 using the \textit{linear} terminal layer. The longitudinal error is quite similar compared with the PREVENTION dataset results. However, the lateral error is exceptionally low, which suggests that the lateral displacements are more reduced in the HigHD dataset.

Table \ref{table:rmse_comparison} shows the performance of different trajectory prediction models reviewed in section \ref{sec:arte} expressed as the RMSE for different prediction horizons.

\begin{table}[ht]
    \renewcommand{\arraystretch}{1.1}
    \centering
    \caption{Quantitative results for the HighD benchmark expressed as RMSE over 3 seconds prediction horizon.}
    \label{table:rmse_comparison}
    \begin{tabular}{ l | c | c | c }
                                   & \multicolumn{3}{c}{Prediction Horizon}\\
    Prediction Horizon             & $t = 1.0$  & $t = 2.0$  & $t = 3.0$ \\
    \hline
    S-LSTM\cite{Alahi2016}         & 0.19 & 0.57 & 1.18 \\
    CS-LSTM\cite{Deo2018_2}        & 0.19 & 0.57 & 1.16 \\
    S-GAN\cite{gupta2018social}    & 0.30 & 0.78 & 1.46 \\
    Multi-head Att.\cite{Kim2020}  & 0.43 & 0.47 & 0.90 \\
    DLM\cite{Khakzar2020}          & 0.22 & 0.61 & 1.16 \\
    PiP\cite{song2020pip}          & 0.17 & 0.52 & 1.05 \\
    CNN+CSP+LSTM\cite{Nejad2021}   & 0.42 & 0.88 & 1.26 \\     
    Ours                  & 0.53 & 0.82 & 1.23 \\
    \end{tabular}
\end{table}

It can be observed that the performance of the proposed model in the early stage of the prediction (1 sec) is not as good as the one achieved by other state-of-the-art methods. This faulty behavior at short prediction horizons could be motivated by the handicap of performing graphic predictions and the associated encoding and decoding procedures. This effect is particularly prominent in data obtained from long distances (e.g. from infrastructure or from a drone) where we have lower longitudinal and lateral accuracy. This gap is narrower for a 2 seconds prediction horizon and the performance is comparable to other methods for 3 seconds.

\section{Conclusions and Future Work}\label{sec:conclusions}

This approach is unlimited in the number of considered vehicles and includes vehicle interactions without any constraints. Context, understood as road configuration is also considered by representing lane markings.
The predictions are generated for all the vehicles in a single step, unlike most of the state-of-the-art works, where a single vehicle is considered as the prediction target and the others actuate as conditioning factors.

The U-net has been selected as prediction core due to its expandable reception field. It can generate a contact area between each single output pixel and a configurable area at the input side.
Several configurations have been explored including different kinds of vehicle representations, context, and network depth levels. Results suggest the deeper the network the better the prediction performance. Gaussian representations of vehicles show better performance compared with rectangular vehicle representations. The representation of the road configuration has no relevant effects on the prediction performance, showing similar results for the experiments conducted with or without lane markings. The best performing configuration has 6 depth levels and uses Gaussian representations of vehicles without lane representations. The prediction error is up to 50\% lower than the baseline method. This system has been trained using on-board sensor data which enables its direct integration on automated vehicles. The HighD has been used as benchmarking dataset to compare the performance of the proposed method with others. Our approach achieves state-of-the-art results on datasets obtained from infrastructure or top-view recording systems and states a baseline for the PREVENTION dataset.

After the review of the state-of-the-art and based on results and conclusions derived from this work, several research lines can be followed to improve the performance of the system or either take advantage of this system in other applications. 
Trajectory prediction results have shown that prediction performance grows with the U-net depth levels. Due to hardware limitations, higher depth levels could not be either trained or tested. More efficient representations will allow trying these deeper models. Hardware improvements will also help to try these models. The general definition of the problem, which exploits visual representations, allows the use of this model in other similar scenarios, such as urban intersections with on-board or external sensor data.

\section*{Acknowledgment}
This work was supported in part by the Spanish Ministry of Science and Innovation under Grants PID2020-114924RB-I00 and DPI2017-90035-R, and in part by the Community Region of Madrid under Grant S2018/EMT-4362 SEGVAUTO 4.0-CM. David Fernández Llorca acknowledges funding from the HUMAINT project at the Digital Economy Unit at the Directorate-General Joint Research Centre (JRC) of the European Commission.

\bibliography{sn-article}


\begin{wrapfigure}{l}{25mm} 
\includegraphics[width=1in,height=1.25in,clip,keepaspectratio]{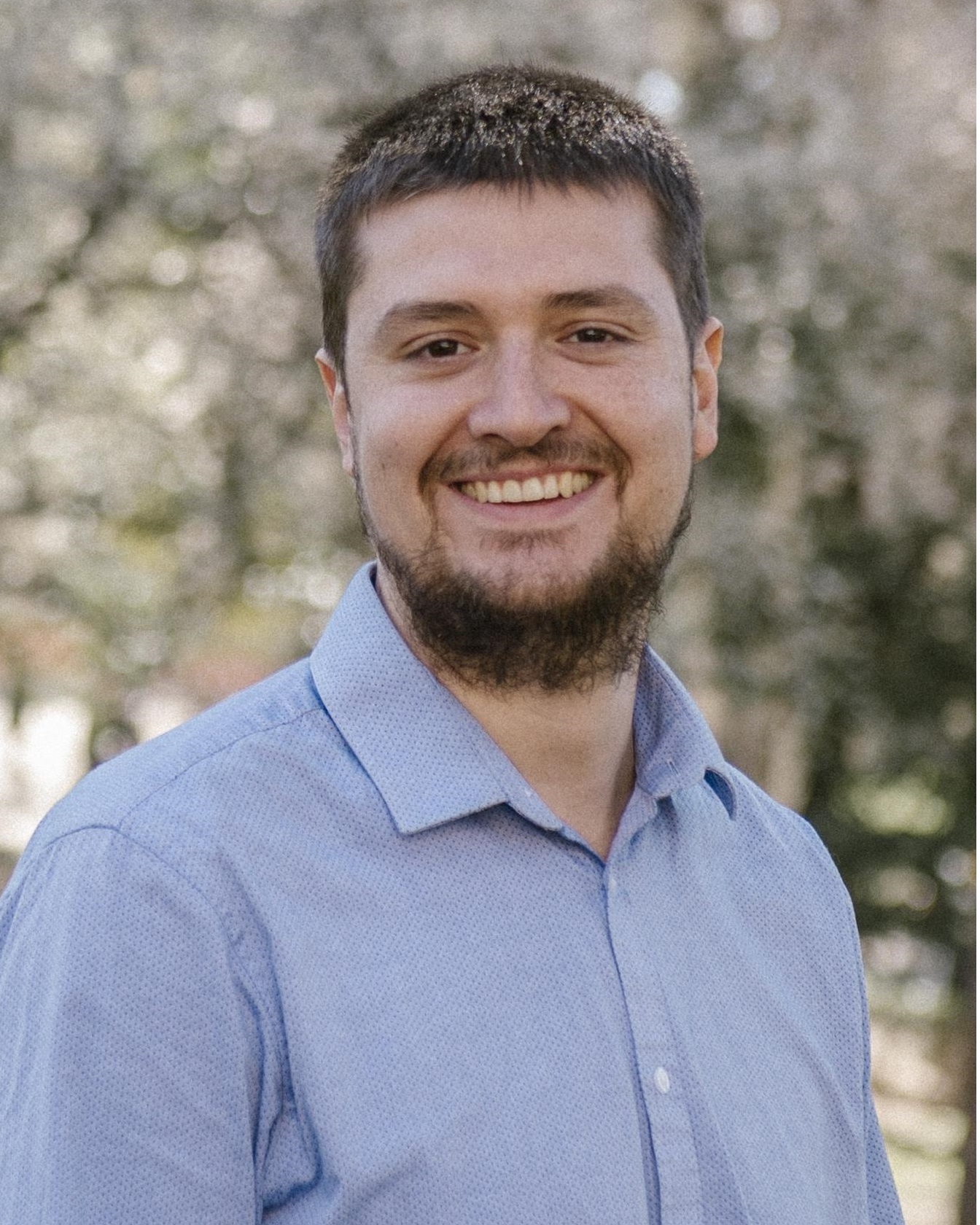}
\end{wrapfigure}\par
\textbf{Rub\'en Izquierdo}  received the Bachelor's degree in electronics and industrial automation engineering in 2014, the M. S. in industrial engineering in 2016, and the Ph.D. degree in information and communication technologies 2020 from the Universidad de Alcalá (UAH). He is currently a Postdoc researcher at the Computer Engineering Department, UAH. His research interest is focused on the prediction of vehicle behaviors but also in control algorithms for highly automated and cooperative vehicles. His work has developed a predictive ACC and AES system for cut-in collision avoidance.\par

\begin{wrapfigure}{l}{25mm} 
\includegraphics[width=1in,height=1.25in,clip,keepaspectratio]{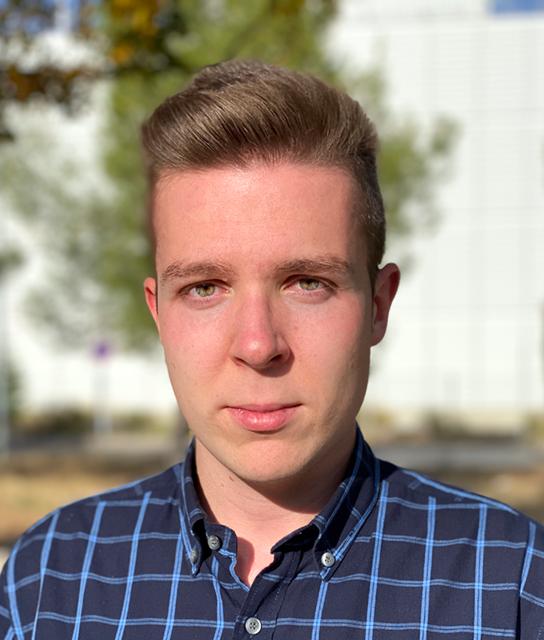}
\end{wrapfigure}\par
\textbf{\'Alvaro Quintanar} (Graduate Student Member, IEEE) obtained the Bachelor's Degree in Telecommunications Engineering in 2017 from Universidad de Alcalá (UAH), specializing in Telecommunication Systems, and the Master's Degree in Telecommunications Engineering in 2019 from Universidad de Alcalá (UAH), earning a specialization in Intelligent Transportation Systems. He started his work in the INVETT Research Group in October 2018, where he is currently pursuing the PhD degree in Information and Communications Technologies with the Computer Engineering Department, developing interactive prediction systems that could forecast trajectories and intention of other road users, such as human-driven vehicles and VRUs.\par

\begin{wrapfigure}{l}{25mm} 
\includegraphics[width=1in,height=1.25in,clip,keepaspectratio]{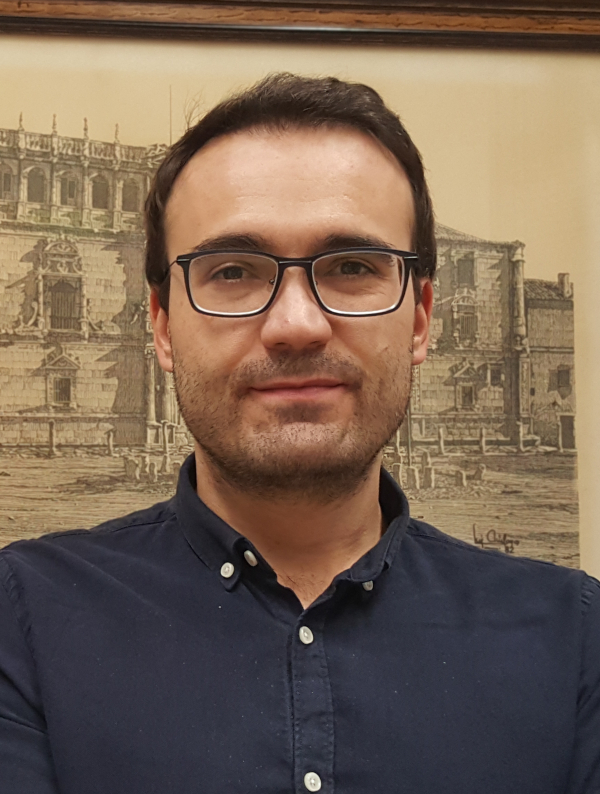}
\end{wrapfigure}\par
\textbf{David Fern\'andez-Llorca} (Senior Member, IEEE) received the Ph.D degree in telecommunication engineering from the University of Alcal\'a (UAH) in 2008. He is currently Scientific Officer at the European Commission - Joint Research Center. He is also Full Professor with UAH. He has authored over 130 publications and more than  10  patents. He received the IEEE ITSS Young Research Award in 2018 and the IEEE ITSS Outstanding Application Award in 2013. He is Editor-in-Chief of the IET Intelligent Transport Systems. His current research interest includes trustworthy AI for transportation, predictive perception for autonomous vehicles, human-vehicle interaction, end-user oriented autonomous vehicles and assistive intelligent transportation systems.\par

\begin{wrapfigure}{l}{25mm} 
\includegraphics[width=1in,height=1.25in,clip,keepaspectratio]{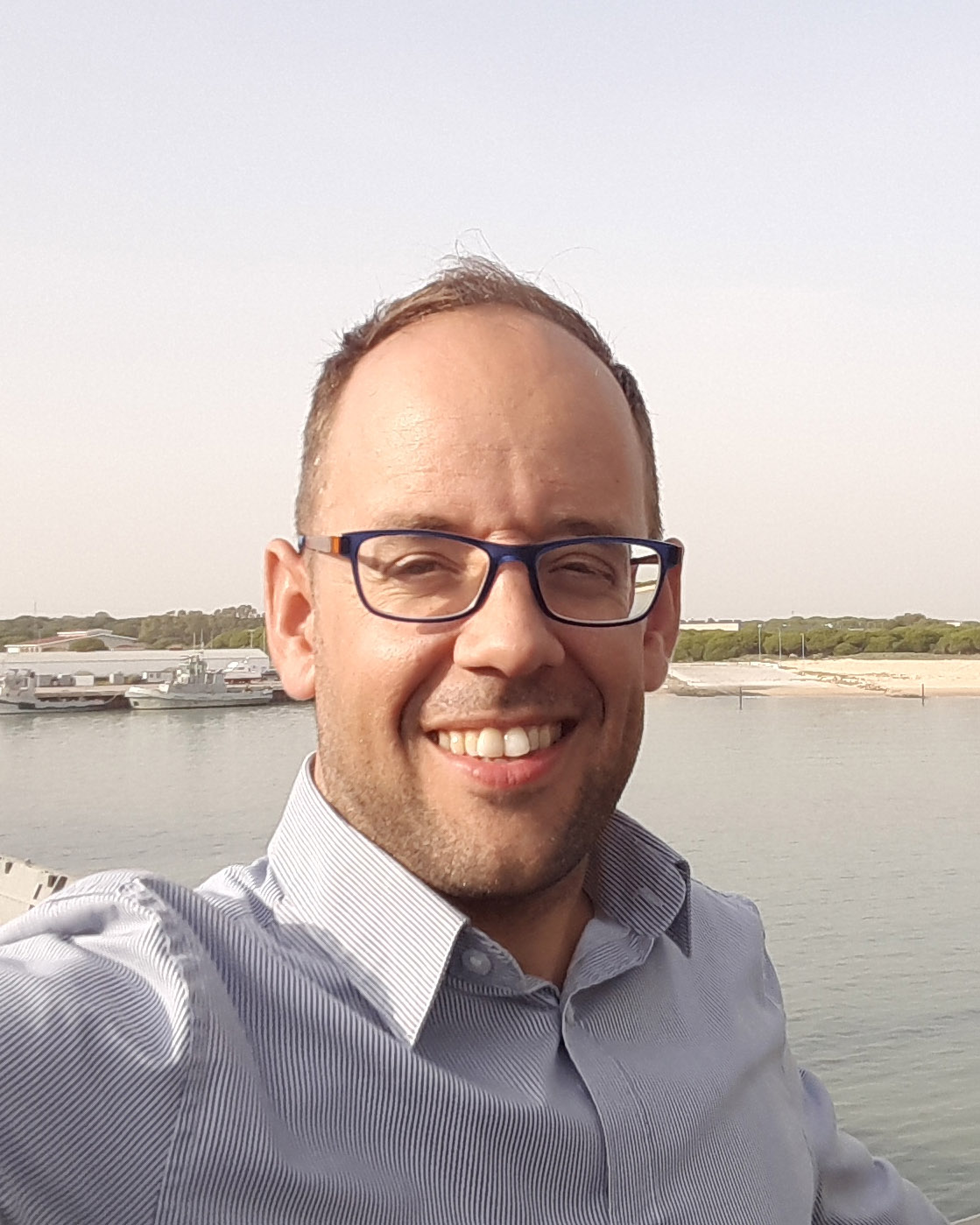}
\end{wrapfigure}\par
\textbf{Iván García-Daza} received the MSc and PhD degrees in Telecommunications Engineering from the University of Alcalá (UAH), Madrid (Spain), in 2004 and 2011 respectively. At present he is Assistant Professor at the Computer Engineering Department at the University of Alcalá and member of the INVETT research group since 2007. In this period he has collaborated on more than 20 projects with public and private funding. All the projects are related with computer science techniques applied on Intelligent Transportation System.\par

\begin{wrapfigure}{l}{25mm} 
\includegraphics[width=1in,height=1.25in,clip,keepaspectratio]{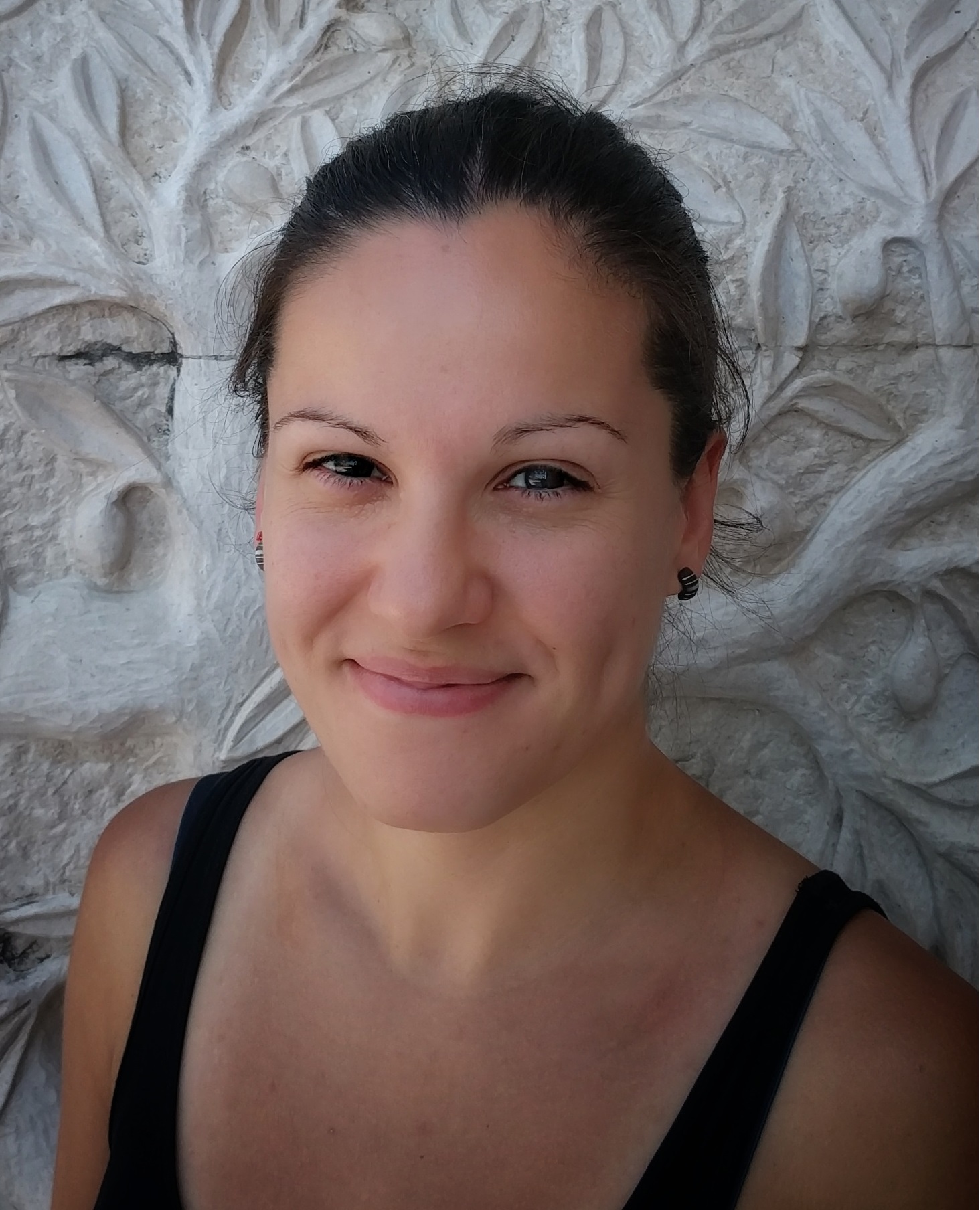}
\end{wrapfigure}\par
\textbf{Noelia Hernández} received the M.Sc. and Ph.D. degrees in advanced electronics systems (intelligent systems) from the University of Alcalá (UAH), in 2009 and 2014, respectively. Her thesis presented a new approach for estimating the global position of a mobile device in indoor environments by using Wi-Fi devices, which received the Best Ph.D. Award by UAH, in 2014. She is currently an Associate Professor with the Computer Engineering Department, UAH. Her research interests include indoor and outdoor localization, artificial intelligence, and intelligent transportation systems.\par

\begin{wrapfigure}{l}{25mm} 
\includegraphics[width=1in,height=1.25in,clip,keepaspectratio]{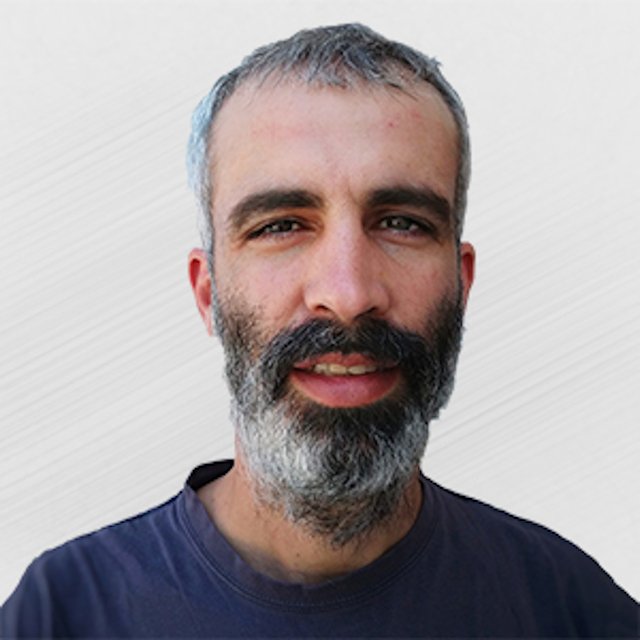}
\end{wrapfigure}\par
\textbf{Ignacio Parra} received the M.S. and Ph.D. degrees in telecommunications engineering from the University of Alcalá (UAH), in 2005 and 2010, respectively. He is currently an Associate Professor with the Computer Engineering Department, UAH. His research interests include intelligent transportation systems and computer vision. He received the Master Thesis Award in eSafety from the ADA Lectureship at the Technical University of Madrid, Spain, in 2006.\par

\begin{wrapfigure}{l}{25mm} 
\includegraphics[width=1in,height=1.25in,clip,keepaspectratio]{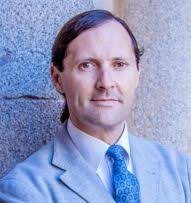}
\end{wrapfigure}\par
\textbf{Miguel \'Angel Sotelo} received the degree in Electrical Engineering in 1996 from the Technical University of Madrid, the Ph.D. degree in Electrical Engineering in 2001 from the University of Alcalá (Alcalá de Henares, Madrid), Spain, and the Master in Business Administration (MBA) from the European Business School in 2008. He is currently a Full Professor at the Department of Computer Engineering of the University of Alcalá. His research interests include Self-driving cars and Predictive Systems. He is author of more than 250 publications in journals, conferences, and book chapters. He has been recipient of the Best Research Award in the domain of Automotive and Vehicle Applications in Spain in 2002 and 2009, and the 3M Foundation Awards in the category of eSafety in 2004 and 2009. Miguel Ángel Sotelo has served as Project Evaluator, Rapporteur, and Reviewer for the European Commission in the field of ICT for Intelligent Vehicles and Cooperative Systems in FP6 and FP7. He is member of the IEEE ITSS Board of Governors and Executive Committee. Miguel Ángel Sotelo served as Editor-in-Chief of the Intelligent Transportation Systems Society Newsletter (2013), Editor-in-Chief of the IEEE Intelligent Transportation Systems Magazine (2014-2016), Associate Editor of IEEE Transactions on Intelligent Transportation Systems (2008-2014), member of the Steering Committee of the IEEE Transactions on Intelligent Vehicles (since 2015), and a member of the Editorial Board of The Open Transportation Journal (2006-2015). He has served as General Chair of the 2012 IEEE Intelligent Vehicles Symposium (IV’2012) that was held in Alcalá de Henares (Spain) in June 2012. He was recipient of the 2010 Outstanding Editorial Service Award for the IEEE Transactions on Intelligent Transportation Systems, the IEEE ITSS Outstanding Application Award in 2013, and the Prize to the Best Team with Full Automation in GCDC 2016. He was President of the IEEE Intelligent Transportation Systems Society (2018-2019).\par

\end{document}